\documentclass[letterpaper]{article} 
\usepackage{aaai2026}  
\usepackage{times}  
\usepackage{helvet}  
\usepackage{courier}  
\usepackage[hyphens]{url}  
\usepackage{graphicx} 
\usepackage{natbib}  
\usepackage{caption} 
\usepackage{algorithm}
\usepackage{algorithmic}

\usepackage{newfloat}
\usepackage{listings}
\DeclareCaptionStyle{ruled}{labelfont=normalfont,labelsep=colon,strut=off} 
\floatstyle{ruled}
\newfloat{listing}{tb}{lst}{}
\floatname{listing}{Listing}
\usepackage{booktabs}
\usepackage{amsfonts}
\usepackage{xcolor}
\usepackage{subcaption}

\usepackage{enumitem}
\usepackage{multirow}
\usepackage{array}
\usepackage{xspace}
\usepackage{makecell}
\usepackage{amsmath}
\usepackage{dsfont}

\usepackage{colortbl}
\usepackage{adjustbox}

\newcommand{\proposal}{{\textit{SkipAlign}}\xspace}
\newcommand{\concept}{{SNA}\xspace}

\title{Let the Void Be Void: Robust Open-Set Semi-Supervised Learning via\\Selective Non-Alignment}
\author{
    You Rim Choi\textsuperscript{\rm 1},
    Subeom Park\textsuperscript{\rm 1},
    Seojun Heo\textsuperscript{\rm 1},
    Eunchung Noh\textsuperscript{\rm 2},
    Hyung-Sin Kim\textsuperscript{\rm 1}
}
\affiliations{

    \textsuperscript{\rm 1}Graduate School of Data Science, Seoul National University, Seoul, South Korea\\
    \textsuperscript{\rm 2} Samsung Electronics\\
    yrchoi@snu.ac.kr, sbpark7@snu.ac.kr, seodalzzz@snu.ac.kr, eunchung.noh@samsung.com, hyungkim@snu.ac.kr
    
}

\begin{document}

\maketitle

\begin{abstract}
Open-set semi-supervised learning (OSSL) leverages unlabeled data containing both in-distribution (ID) and unknown out-of-distribution (OOD) samples, aiming simultaneously to improve closed-set accuracy and detect novel OOD instances. Existing methods either discard valuable information from uncertain samples or force-align every unlabeled sample into one or a few synthetic ``catch-all'' representations, resulting in geometric collapse and overconfidence on only seen OODs.
To address the limitations, we introduce \textit{selective non-alignment}, adding a novel ``skip'' operator into conventional pull and push operations of contrastive learning. Our framework, \proposal, selectively skips alignment (pulling) for low-confidence unlabeled samples, retaining only gentle repulsion against ID prototypes. This approach transforms uncertain samples into a pure repulsion signal, resulting in tighter ID clusters and naturally dispersed OOD features.
Extensive experiments demonstrate that \proposal significantly outperforms state-of-the-art methods in detecting unseen OOD data without sacrificing ID classification accuracy.

\end{abstract}    
\section{Introduction}
\label{sec:intro}

\begin{figure*}[t]
  \centering
    \includegraphics[width=.92\linewidth]{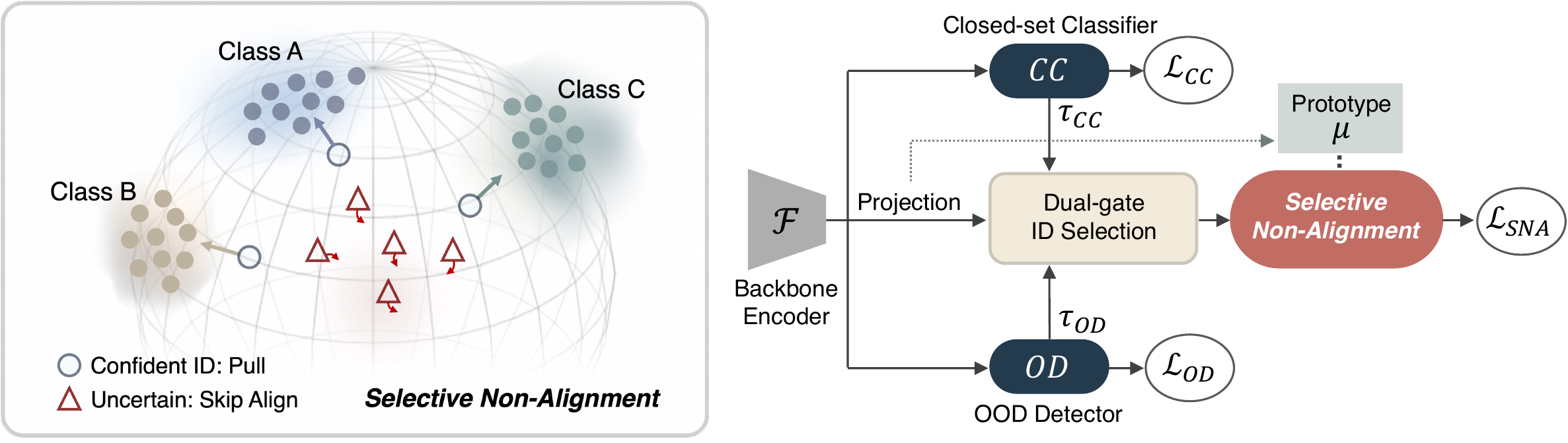}
    \caption{
    Overview of \proposal. 
    Left: Illustration of the Selective Non-Alignment (SNA) concept. Confident ID samples are attracted toward their corresponding class prototypes, while uncertain or OOD samples are softly repelled and remain unaligned.  
    Right: The overall framework of \proposal. A shared encoder feeds into a closed-set classifier ($CC(\cdot)$) and an OOD detector ($OD(\cdot)$), which jointly perform dual-gate ID selection. Projection embeddings are used to form class prototypes, enabling SNA to refine representation learning through prototype-based alignment and repulsion.
    }
  \label{fig:overview}
\end{figure*}

Semi-supervised learning (SSL) improves performance by leveraging unlabeled data, typically assuming that \textit{all} unlabeled samples belong strictly to one of the known in-distribution (ID) classes~\cite{sohn2020fixmatch}. However, practical unlabeled datasets inevitably form \textbf{open sets}, containing unknown out-of-distribution (OOD) samples. To address this limitation, Open-Set Semi-Supervised Learning (OSSL) learns from open-set unlabeled data to simultaneously improve ID classification and reliably detect OOD samples, 
including OOD classes present in the unlabeled pool (\textit{seen} OOD) and novel OOD classes that appear only at test time (\textit{unseen} OOD).

However, generalizable OOD detection via OSSL remains challenging. This is primarily because \textbf{the true OOD space is inherently vast and heterogeneous, whereas training provides only limited, potentially misleading glimpses of OOD samples.} 
Recent state-of-the-art methods take opposite stances toward OOD: aggressive or conservative. 
On the aggressive side, SCOMatch~\cite{wang2024scomatch} collapses the diverse OOD universe into a single synthetic class, causing \textit{geometric collapse and overconfidence on seen OOD samples}. 
Similarly, ProSub~\cite{wallin2024prosub} creates a broad ID subspace spanning all ID clusters, aggressively pulling or pushing every unlabeled sample toward or away from this subspace even when pseudo-labels are \textit{noisy}. Consequently, unseen OOD samples that \textit{fall within the broad ID subspace} but outside any specific ID cluster tend to be misclassified as ID.
On the conservative side, SSB~\cite{fan2023ssb} trains per-class one-vs-all (OVA) classifiers but discards low-confidence unlabeled data, \textit{losing valuable geometric cues} that could improve feature space discrimination between ID and novel OOD.

To address these limitations, we conceptualize the embedding space as a universe consisting of compact \textbf{galaxies} (ID classes) surrounded by a vast \textbf{interstellar void} (OOD). Guided by this analogy, we propose \textit{Selective Non-Alignment (\concept)}, a novel contrastive learning principle that adds a third operator---``\textbf{skip}''---alongside traditional pull-push pair. 
\concept pursues two objectives: (1) preserve the natural heterogeneity of OOD samples, and (2) carefully guide uncertain unlabeled samples to minimize the risk of amplifying ID/OOD pseudo-labeling errors.

Specifically, an unlabeled sample is pulled into an ID galaxy only when it passes a \textbf{stringent confidence test}, significantly reducing the risk of mistakenly aligning an OOD sample.
Otherwise, unlike previous prototype‑based contrastive methods that attract every sample toward a prototype~\cite{li2020prototypical}, \concept \textbf{nullifies the pull term} and applies only a \textit{non‑selective angular update} computed from a softmax‑weighted average of all ID prototype directions. This ensures that points closer to ID galaxies experience stronger repulsion, preventing premature attachment to any galaxy. 
Over time, genuine ID samples in the void naturally gravitate toward their correct galaxies as the model matures, whereas genuine OODs remain securely dispersed. This produces a distinctive geometric structure for robust detection of unseen OODs: \textbf{dense ID galaxies clearly separated by a diffuse void}. Figure~\ref{fig:overview} illustrates the concept of \concept.

We instantiate \concept in the \proposal framework, integrating it alongside an ID classifier and an OVA OOD detector sharing a unified backbone. To ensure representational independence and flexibility, \proposal employs an additional non-linear projection layer before applying \concept. 
During training, SNA synergistically interacts with the ID classifier and OOD detector in two ways: (1) For the stringent confidence criterion of SNA, \proposal utilizes a dual-gate mechanism, requiring simultaneous high-confidence agreement from both the ID classifier and OOD detector, and (2) SNA progressively refines the shared backbone to enhance performance on both downstream tasks.
Furthermore, \proposal progressively refines prototypes  by integrating confidently pseudo-labeled unlabeled samples into class centroids, stabilizing training. 

Our contributions can be summarized as follows:
\begin{itemize}[leftmargin=*,noitemsep,topsep=0pt]
    \item \textbf{Selective non-alignment principle:} We introduce \textit{skip} as an essential third contrastive operator, effectively preventing OOD overfitting by selectively suppressing alignment.

    \item \textbf{Tailored framework integration:} \proposal seamlessly integrates \concept with a dedicated projection layer, dual-gate confidence checks, and progressive prototype refinement to robustly leverage both labeled and unlabeled data.

    \item \textbf{Extensive empirical validation:}
    Extensive evaluations on CIFAR-10/100, ImageNet-30, and TinyImageNet, and diverse unseen OOD datasets demonstrate \proposal's superior generalization, improving Overall-OOD AUC by +3.1 points on average (max +7.1) over the strongest baseline, while maintaining comparable closed‑set accuracy. 
\end{itemize}

\section{Related Work}
\label{sec:related_work}

\subsection{Open-set Semi-Supervised Learning (OSSL)}
 
Early approaches to OSSL identify and exclude OOD samples, thus minimizing their negative impact on ID classifiers~\cite{yu2020multi, guo2020safe, saito2021openmatch}. 
Recent methods, however, have shifted toward actively incorporating OOD samples. To leverage representations from OOD data, auxiliary self-supervised tasks, such as orientation prediction~\cite{huang2021trash} or feature consistency enforcement~\cite{wallin2024improving}, have been explored. However, these general-purpose tasks do not directly target OOD detection, resulting in limited gains.

More recent, targeted OSSL approaches handle OOD samples either aggressively or conservatively. 
On the aggressive side, IOMatch~\cite{li2023iomatch} and SCOMatch~\cite{wang2024scomatch}  introduce OSSL-specific modules using a $(K + 1)$-way classifier, assigning a single additional class for all OOD samples. While SCOMatch improves over IOMatch by enhancing pseudo-label reliability, collapsing diverse OOD samples into one homogeneous class disrupts the embedding space geometry and induces overfitting to seen OOD samples, thereby hindering generalization to novel OOD data. 
Similarly, ProSub~\cite{wallin2024prosub} defines an ID subspace  spanning class prototypes and actively repels OOD samples from the subspace based on angular distances, establishing clear boundaries for known OODs. Despite being less rigid than the single-class strategy, ProSub's subspace assumption can inadvertently classify unseen OOD samples located within the ID subspace but outside specific ID prototypes as ID, leading to detection errors. 
Furthermore, these aggressive methods consistently align unlabeled samples to prototypes or classes regardless of pseudo-label noise, compromising performance due to incorrect alignments.

On the conservative side, SSB~\cite{fan2023ssb} employs one-vs-all (OVA) binary classifiers for each ID class---partitioning samples into ``ONE'' (target ID class) and ``ALL'' (other classes) categories---with the OOD region defined as the intersection of all ``ALL'' regions. During training, the ``ONE'' class exclusively uses labeled samples, while the ``ALL'' class additionally includes only a limited subset of highly confident OOD samples. Although this conservative approach reduces the risk of erroneous alignment, discarding low-confidence unlabeled data sacrifices valuable geometric cues essential for distinguishing ID from novel OOD.

Therefore, achieving robust and generalizable OOD detection through OSSL remains an open challenge, demanding (1) a tailored characterization of the OOD space and (2) a nuanced approach to handling uncertain unlabeled samples.

\section{\proposal}

We propose \proposal to improve generalizable OOD detection in OSSL by addressing the key limitations of existing methods.  
Our core innovation for OSSL is \textbf{Selective Non-Alignment (SNA)}, characterized by two complementary strategies.
First, SNA forms compact, individual ID clusters (no subspace spanning), selectively pulling only high-confidence ID samples toward their respective class prototypes. 
Second, for all other unlabeled samples---including confidently identified OOD and uncertain samples---SNA completely removes the pull operation, applying instead a non-selective angular repulsion. This has two benefits: (1) By refraining from pulling confident OOD samples to specific regions, \proposal preserves the inherent diversity and dispersion of the OOD interstellar void; (2) Mildly pushing uncertain samples from all prototypes provides robust contrastive learning that effectively mitigates sensitivity to pseudo-label noise.


\subsection{Overview of \proposal}
For $\mathit{K}$-way classification, let $ \mathit{X} = \{(x_i, y_i)\}_{i=1}^{B}$ be a batch of $\mathit{B}$ labeled samples from $\mathcal{D}_l$ with labels $y_i \in \{1, ..., K\}$.  
Let $\mathcal{U} = \{u_i\}_{i=1}^{\gamma B}$ be $\gamma B$ unlabeled samples from $\mathcal{D}_u$, where $\gamma$ controls the labeled-to-unlabeled ratio.
Unlike SSL, OSSL includes \textit{OOD samples} from unknown classes in $\mathcal{D}_{u}$.

As illustrated in Figure~\ref{fig:overview}, \proposal consists of a shared backbone and three task-specific heads. These heads share representations and mutually reinforce learning---particularly the SNA module aiding both classification and OOD detection---while MLP projection heads are inserted to reduce harmful gradient interference, preserving constructive information flow across tasks.
Each head is optimized for a specific objective:  
\begin{itemize}[leftmargin=*,noitemsep,topsep=0pt]
\item \textbf{Closed-set (ID) classifier} ($CC$) performs standard $\mathit{K}$-way classification over ID classes, outputting a softmax probability vector $\mathbf{p} \in \mathbb{R}^{\mathit{K}}$.
\item \textbf{OOD detector} ($OD$) distinguishes ID from OOD samples using an OVA classifier with $\mathit{K}$ binary sub-classifiers. 
Each detector $OD_k(\cdot)$ outputs a probability pair $(\varphi_{k}^{ID}, \varphi_{k}^{OOD})$ indicating the likelihood of being ID or OOD for class $k$.
\item \textbf{Selective Non-Alignment (SNA) module}, the core of \proposal, shapes the feature space to improve both ID classification and OOD detection, facilitating better generalization to unseen OODs. It operates on the embedding $\mathbf{z} \in \mathbb{R}^{d}$ produced by a dedicated projection head on top of the shared backbone.
\end{itemize}


\subsection{Selective Non-Alignment}

Unlabeled data in OSSL inevitably include samples from unknown classes, introducing ambiguity in their assignment to ID prototypes.
To address this, the SNA module attracts only those samples confidently predicted as ID by both the closed-set classifier and the OOD detector—a mechanism we term \textbf{dual-gate ID selection}.
For each unlabeled embedding $z_i$, its confidence mask $\Phi_i$ is defined as:
\begin{equation} \label{eqn:eq1}
\Phi_{i} = \mathds{1}\big( p_{i,\hat{k}} > \tau_{\scriptscriptstyle ID} \land \varphi_{i,\hat{k}}^{ID} > \eta_{\scriptscriptstyle ID} \big),
\quad \hat{k} = \arg\max_k p_{i,k}
\end{equation}
where $\tau_{\scriptscriptstyle ID}$ and $\eta_{\scriptscriptstyle ID}$ are thresholds for the closed-set classifier and the OOD detector, respectively.

Let $\mathrm{sim}(a,b)=\frac{a}{\|a\|}\cdot\frac{b}{\|b\|}$ denote the cosine similarity between two vectors. 
For an unlabeled embedding $z_i$ with predicted class $\hat{k}$ and its prototype $\mu_{\hat k} \in \mathbb{R}^d$, the unlabeled SNA loss is:
\begin{equation}\label{eqn:eq2}
\mathcal{L}_{\text{USNA}}(z_i)
= -\Phi_i\,\frac{\mathrm{sim}(z_i,\mu_{\hat k})}{T}
+\log\sum_{j=1}^{K}\exp\Big(\frac{\mathrm{sim}(z_i,\mu_j)}{T}\Big)
\end{equation}
where $T$ is a temperature parameter.

The gradient of $\mathcal{L}_{\text{USNA}}$ with respect to $z_i$ is:
\begin{equation}\label{eq:grad_z}
\nabla_{z_i}\mathcal L_{\text{USNA}}
=\frac{1}{T\|z_i\|}\big(I-\hat z\hat z^{\top}\big)\Big(\sum_{j=1}^K \alpha_j\,\hat\mu_j-\Phi_i\,\hat\mu_{\hat k}\Big)
\end{equation}
where $\hat z = z_i/\|z_i\|$, $\hat\mu_j=\mu_j/\|\mu_j\|$, and 
$\alpha_j$ is the softmax probability of $z_i$ belonging to class $j$ ($\frac{\exp(\mathrm{sim}(z_i,\mu_j)/T)}{\sum_\ell \exp(\mathrm{sim}(z_i,\mu_\ell)/T)}$). 
The projection term $(I-\hat z\hat z^\top)$ removes radial components, ensuring that updates act purely in the angular direction.

This design selectively aligns only confident ID samples ($\Phi_i=1$) toward their respective class prototypes. In contrast, uncertain or OOD samples ($\Phi_i=0$) receive \textbf{no targeted pull and instead undergo a non-selective angular update} determined by the averaged term $\sum_j \alpha_j\hat\mu_j$, preventing unwanted alignment to any specific prototype. 
Complementing the $CC$ and $OD$ heads, \concept addresses a key limitation of their objectives (e.g., cross-entropy or pseudo-label consistency), which may unintentionally enlarge feature norms even for incorrectly pseudo-labeled samples. 
\concept's stricter dual-gate mechanism classifies these ambiguous samples as uncertain ($\Phi_i=0$). The corresponding $\mathcal{L}_{\text{USNA}}$ then provides an orthogonal corrective ``torque'', redirecting uncertain features away from incorrectly matched prototypes without aligning them elsewhere. Consequently, norm growth occurs exclusively along reliable directions, effectively mitigating the risk of
uncertain samples from accumulating large norms in misleading regions. 
Jointly, \concept, $CC$, and $OD$, yield a more discriminative and robust feature space, suitable for challenging open-set scenarios.

For labeled samples, we exploit reliable ground-truth supervision to promote intra-class compactness and implicit inter-class separation.
The Instance-wise Alignment loss ($\mathcal{L}_{\text{IA}}$), following the supervised contrastive formulation~\cite{khosla2020supervised}, attracts same-class samples:
\begin{equation} \label{eqn:eq3}
\mathcal{L}_{\text{IA}}(z_i) = -\frac{1}{|P(i)|} \sum_{p \in P(i)} \log \frac{\exp(\mathrm{sim}(z_i, z_p) / T)}{\sum_{j=1}^{\mathit{B}} \mathds{1}_{i \neq j} \exp(z_i \cdot z_j / T)}
\end{equation}
where $P(i)$ is the set of same-class samples excluding $i$, and $T$ is the temperature.
Prototype-based Alignment loss ($\mathcal{L}_{\text{PA}}$) always pulls each embedding toward its class prototype:
\begin{equation}\label{eqn:eq4}
\mathcal{L}_{\text{PA}}(z_i)
= -\frac{\mathrm{sim}(z_i,\mu_{y_i})}{T}
+\log\sum_{j=1}^{K}\exp\Big(\frac{\mathrm{sim}(z_i,\mu_j)}{T}\Big)
\end{equation}
where $\mu_{y_i}$ denotes the prototype of class $y_i$.

The overall SNA loss combines labeled and unlabeled objectives:
\begin{equation}
\mathcal{L}_{\text{SNA}} = \lambda_{\text{USNA}} \mathcal{L}_{\text{USNA}}(\mathcal{U}) 
+ \lambda_{\text{IA}} \mathcal{L}_{\text{IA}}(\mathit{X}) 
+ \lambda_{\text{PA}} \mathcal{L}_{\text{PA}}(\mathit{X}),
\end{equation}
where $\lambda_{\text{USNA}}, \lambda_{\text{IA}}, \lambda_{\text{PA}}$ weight the contribution of each term.

\begin{figure}[t]
  \centering
    \includegraphics[width=\linewidth]{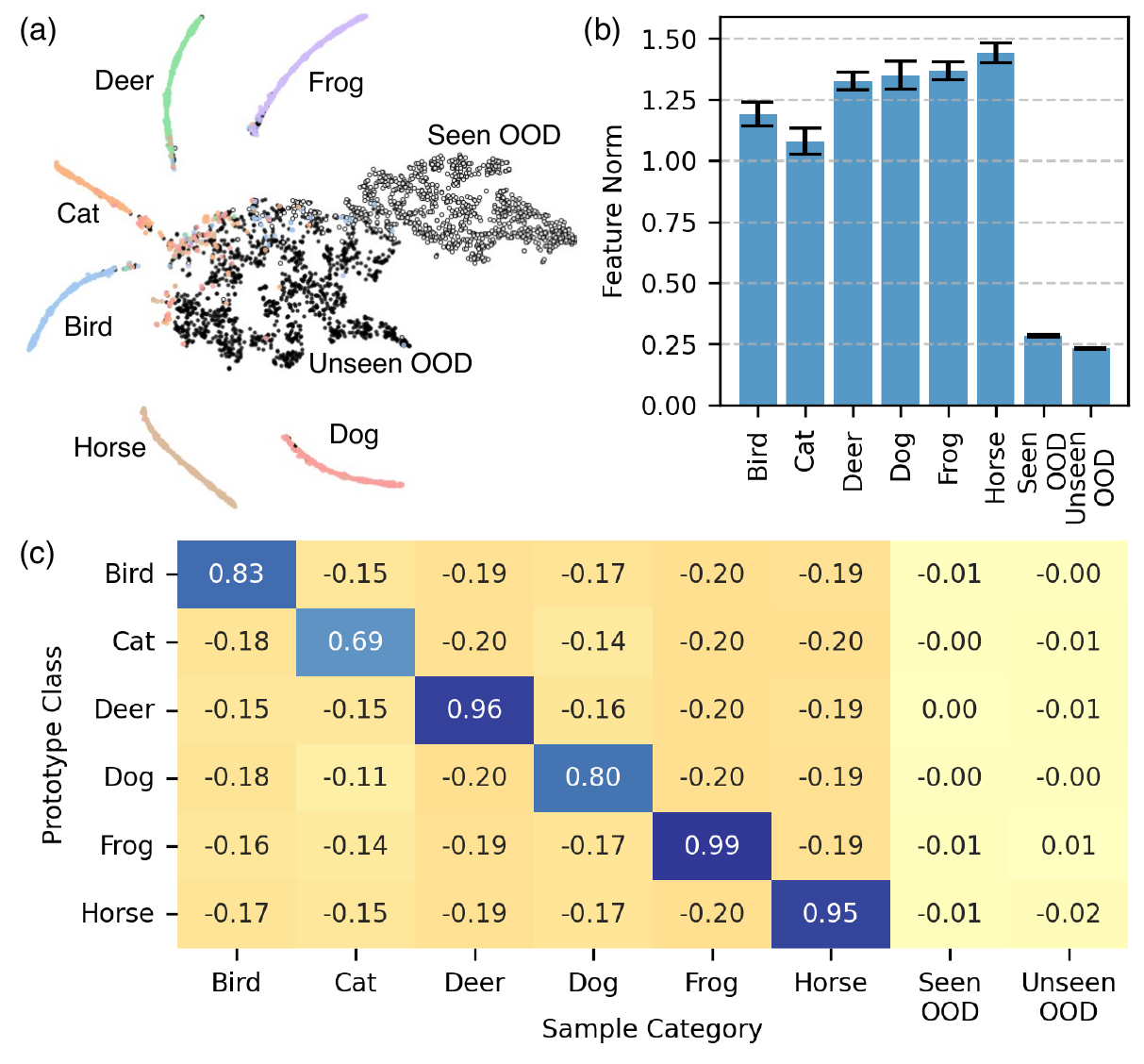}
    \caption{
    Effect of the SNA module.
    (a) t-SNE visualization of projection embeddings $\mathbf{z}$.  
    (b) Average feature norm for each class category.  
    (c) Average cosine similarity between sample categories and ID class prototypes. 
    }
  \label{fig:sna}
\end{figure}

%
Figure~\ref{fig:sna} provides a visual illustration of the geometric effects induced by our SNA module.
This figure is generated under a CIFAR-10 OSSL setting, where six animal classes are treated as inliers with 50 labeled samples per class, four classes as seen OOD, and \textbf{six additional datasets as unseen OOD} (see Section Experiments for details).
Figure~\ref{fig:sna}(a) shows a t-SNE plot of projection embeddings $\mathbf{z}$. 
As intended, ID samples form distinct, compact clusters with smoothly curved manifolds, demonstrating clear alignment with prototype direction. 
In contrast, both seen and unseen OOD samples remain in the surrounding void without attaching to any ID cluster, reflecting the effect of angular updates from SNA.
Some uncertain ID samples, excluded by our dual-gate criterion, lie near the ID clusters but partially intermingle with nearby OOD embeddings.

To further assess discriminability, we analyze feature norms and prototype alignment.
Figure~\ref{fig:sna}(b) reveals a clear norm gap between ground-truth ID and OOD samples.
This separation reflects how USNA encourages reliable samples to align with prototypes—thereby accumulating norm through the synergistic pull of $CC$, $OD$, and SNA alignment—while applying a selective repulsion to misaligned or unlabeled samples.
As a result, ID samples exhibit significantly higher norms than both seen and unseen OODs.
Figure~\ref{fig:sna}(c) illustrates the average cosine similarity between each sample category and the ID prototypes. ID samples exhibit high similarity with their respective prototypes, whereas both seen and unseen OODs consistently display low similarity to all prototypes. 
Collectively, SNA, together with $CC$ and $OD$, naturally shapes a geometric structure consisting of dense, high-norm ID ``galaxies'' clearly separated by a sparse, low-norm OOD ``interstellar void'', effectively generalizing to unseen OOD distributions.


\subsection{Prototype Refinement}
\label{sec:proto_gen}

\begin{figure}[t]
  \centering
    \includegraphics[width=\linewidth]{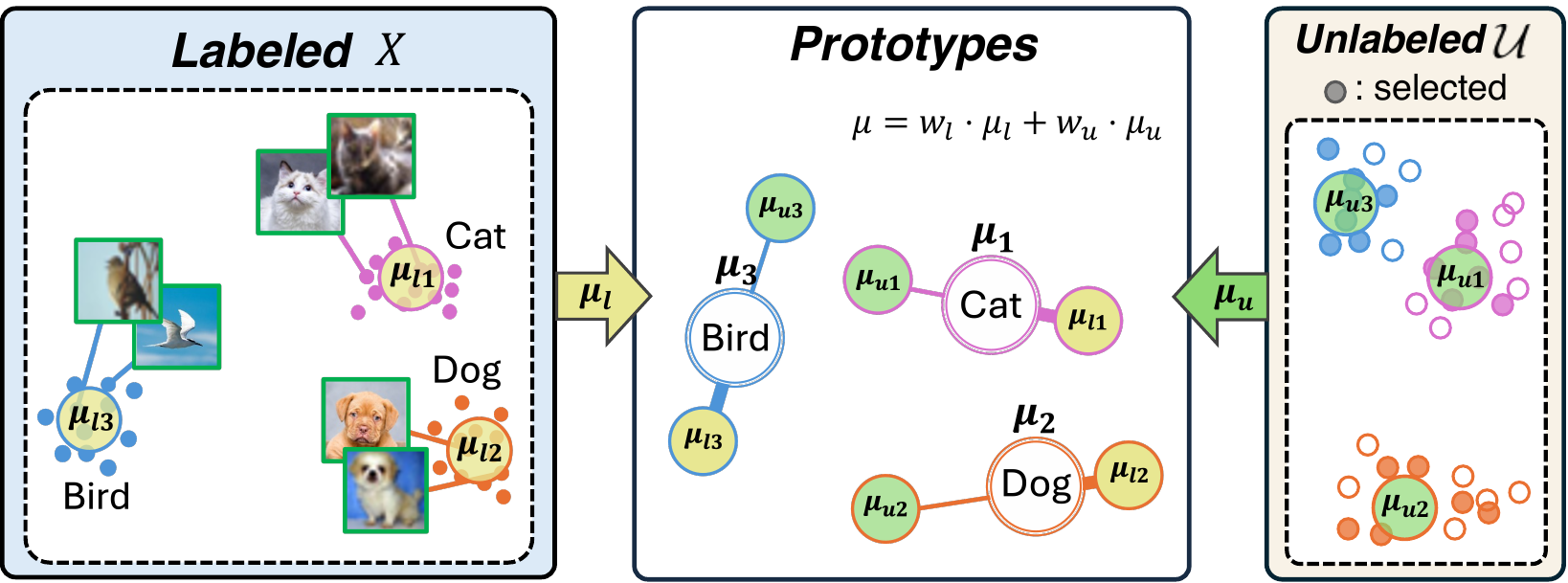}
    \caption{Adaptive Prototype Generation. For each class, the labeled prototype $\mu_l$ is the mean embedding of labeled samples, while the unlabeled prototype $\mu_u$ is obtained from confident ID samples among the unlabeled data. The final prototype $\mu$ is computed as a weighted sum of the two.}
  \label{fig:fig3}
\end{figure}

In \concept, prototypes serve as positive anchors that guide ID embeddings toward class centers. 
However, relying solely on limited labeled data can lead to suboptimal prototypes. 
To mitigate this, we dynamically refine prototypes by combining labeled features with reliably pseudo-labeled unlabeled features selected via a stricter dual-gate criterion. 
Figure~\ref{fig:fig3} illustrates this process, where the final prototype is computed as a weighted average of labeled and selected unlabeled embeddings. 
The weights adaptively account for sample counts and a hyperparameter controlling the influence of unlabeled data, as detailed in Supplementary Material. 
This refinement stabilizes prototype representations and improves class separability throughout training.

\begin{table*}[t]
\centering
\renewcommand{\arraystretch}{0.7}
\setlength{\tabcolsep}{5pt} 
\fontsize{9}{11}\selectfont

\begin{tabular}{c|cc|cc|cc|cc|cc|cc|cc|cc}
\toprule
Dataset & 
\multicolumn{4}{c}{\textbf{CIFAR-10 6/4}} & 
\multicolumn{4}{c}{\textbf{CIFAR-100 55/45}} &
\multicolumn{4}{c}{\textbf{CIFAR-100 80/20}} &
\multicolumn{2}{c}{\textbf{ImageNet30}} & 
\multicolumn{2}{c}{\textbf{TIN}}\\ 
\cmidrule(lr){2-5} \cmidrule(lr){6-9} \cmidrule(lr){10-13} \cmidrule(lr){14-17}
Labels & 
\multicolumn{2}{c}{25} & \multicolumn{2}{c|}{50} & 
\multicolumn{2}{c}{25} & \multicolumn{2}{c|}{50} & 
\multicolumn{2}{c}{25} & \multicolumn{2}{c|}{50} &
\multicolumn{2}{c}{5\%} & \multicolumn{2}{c}{5\%} \\
\cmidrule(lr){2-3} \cmidrule(lr){4-5} \cmidrule(lr){6-7} 
\cmidrule(lr){8-9} \cmidrule(lr){10-11} \cmidrule(lr){12-13} \cmidrule(lr){14-15} \cmidrule(lr){16-17}
Metric & Acc. & AUC & Acc. & AUC & Acc. & AUC &
Acc. & AUC & Acc. & AUC & Acc. & AUC & Acc. & AUC & Acc. & AUC\\

\midrule
\shortstack{IOMatch \\ \scriptsize{}} & 
\shortstack{\textbf{93.2} \\ {\scriptsize $\pm$0.6}} & \shortstack{83.7 \\ {\scriptsize $\pm$0.6}} &
\shortstack{\textbf{93.3} \\ {\scriptsize $\pm$0.4}} & \shortstack{84.9 \\ {\scriptsize $\pm$1.0}} &
\shortstack{69.5 \\ {\scriptsize $\pm$0.5}} & \shortstack{73.6 \\ {\scriptsize $\pm$0.2}} &
\shortstack{73.2 \\ {\scriptsize $\pm$0.0}} & \shortstack{76.6 \\ {\scriptsize $\pm$2.0}} &
\shortstack{\underline{64.5} \\ {\scriptsize $\pm$0.5}} & \shortstack{70.0 \\ {\scriptsize $\pm$1.2}} &
\shortstack{68.6 \\ {\scriptsize $\pm$0.1}} & \shortstack{71.6 \\ {\scriptsize $\pm$1.4}} &
\shortstack{84.5 \\ {\scriptsize $\pm$0.3}} & \shortstack{71.4 \\ {\scriptsize $\pm$2.4}} &
\shortstack{57.9 \\ {\scriptsize $\pm$0.3}} & \shortstack{69.0 \\ {\scriptsize $\pm$1.6}}\\

\shortstack{SSB \\ \scriptsize{}} & 
\shortstack{84.5 \\ {\scriptsize $\pm$3.9}} & \shortstack{75.5 \\ {\scriptsize $\pm$4.5}} &
\shortstack{91.0 \\ {\scriptsize $\pm$1.3}} & \shortstack{\underline{93.5} \\ {\scriptsize $\pm$0.5}} &
\shortstack{69.4 \\ {\scriptsize $\pm$0.8}} & \shortstack{\underline{80.8} \\ {\scriptsize $\pm$0.7}} &
\shortstack{74.0 \\ {\scriptsize $\pm$0.5}} & \shortstack{\underline{83.4} \\ {\scriptsize $\pm$1.7}} &
\shortstack{64.3 \\ {\scriptsize $\pm$0.4}} & \shortstack{\underline{75.7} \\ {\scriptsize $\pm$2.7}} &
\shortstack{\underline{69.8} \\ {\scriptsize $\pm$0.3}} & \shortstack{75.3 \\ {\scriptsize $\pm$1.2}} &
\shortstack{91.8 \\ {\scriptsize $\pm$0.0}} & \shortstack{82.0 \\ {\scriptsize $\pm$0.3}} &
\shortstack{59.9 \\ {\scriptsize $\pm$0.3}} & \shortstack{73.8 \\ {\scriptsize $\pm$1.0}}\\

\shortstack{SCOMatch \\ \scriptsize{}} & 
\shortstack{\underline{92.1} \\ {\scriptsize $\pm$0.6}} & \shortstack{72.0 \\ {\scriptsize $\pm$1.0}} &
\shortstack{92.4 \\ {\scriptsize $\pm$0.3}} & \shortstack{73.3 \\ {\scriptsize $\pm$0.7}} &
\shortstack{\underline{71.0} \\ {\scriptsize $\pm$0.8}} & \shortstack{62.2 \\ {\scriptsize $\pm$0.9}} &
\shortstack{\underline{74.6} \\ {\scriptsize $\pm$0.4}} & \shortstack{60.7 \\ {\scriptsize $\pm$0.6}} &
\shortstack{64.1 \\ {\scriptsize $\pm$0.3}} & \shortstack{65.2 \\ {\scriptsize $\pm$3.0}} &
\shortstack{69.7 \\ {\scriptsize $\pm$0.5}} & \shortstack{72.5 \\ {\scriptsize $\pm$0.8}} & 
\shortstack{87.6 \\ {\scriptsize $\pm$0.5}} & \shortstack{61.4 \\ {\scriptsize $\pm$1.8}} &
\shortstack{51.3 \\ {\scriptsize $\pm$0.4}} & \shortstack{45.8 \\ {\scriptsize $\pm$4.2}}\\

\shortstack{ProSub \\ \scriptsize{}} & 
\shortstack{88.0 \\ {\scriptsize $\pm$1.0}} & \shortstack{\underline{87.8} \\ {\scriptsize $\pm$1.1}} &
\shortstack{88.5 \\ {\scriptsize $\pm$2.1}} & \shortstack{88.0 \\ {\scriptsize $\pm$2.5}} &
\shortstack{63.6 \\ {\scriptsize $\pm$1.0}} & \shortstack{76.5 \\ {\scriptsize $\pm$2.2}} &
\shortstack{69.2 \\ {\scriptsize $\pm$1.4}} & \shortstack{79.2 \\ {\scriptsize $\pm$0.9}} &
\shortstack{55.2 \\ {\scriptsize $\pm$1.1}} & \shortstack{74.6 \\ {\scriptsize $\pm$1.9}} &
\shortstack{61.1 \\ {\scriptsize $\pm$1.9}} & \shortstack{\underline{78.5} \\ {\scriptsize $\pm$3.4}} & 
\shortstack{\underline{92.7} \\ {\scriptsize $\pm$1.0}} & \shortstack{\underline{88.5} \\ {\scriptsize $\pm$0.7}} &
\shortstack{\textbf{61.0} \\ {\scriptsize $\pm$0.2}} & \shortstack{\underline{75.8} \\ {\scriptsize $\pm$0.3}}\\

\midrule
\shortstack{\textbf{\proposal} \\ \scriptsize{}} & 
\shortstack{\rule{0pt}{2ex} 92.0 \\ {\scriptsize $\pm$0.4}} & \shortstack{\textbf{94.9} \\ {\scriptsize $\pm$0.7}} &
\shortstack{\underline{93.1} \\ {\scriptsize $\pm$0.5}} & \shortstack{\textbf{96.6} \\ {\scriptsize $\pm$0.2}} &
\shortstack{\textbf{71.6} \\ {\scriptsize $\pm$0.4}} & \shortstack{\textbf{82.8} \\ {\scriptsize $\pm$1.6}} &
\shortstack{\textbf{75.0} \\ {\scriptsize $\pm$0.2}} & \shortstack{\textbf{83.9} \\ {\scriptsize $\pm$3.2}} &
\shortstack{\textbf{67.2} \\ {\scriptsize $\pm$0.1}} & \shortstack{\textbf{80.1} \\ {\scriptsize $\pm$0.7}} &
\shortstack{\textbf{70.4} \\ {\scriptsize $\pm$0.5}} & \shortstack{\textbf{82.8} \\ {\scriptsize $\pm$0.6}} & 
\shortstack{\textbf{92.8} \\ {\scriptsize $\pm$ 0.4}} & \shortstack{\textbf{90.7} \\ {\scriptsize $\pm$ 1.0}} &
\shortstack{\underline{60.2} \\ {\scriptsize $\pm$0.4}} & \shortstack{\textbf{77.1} \\ {\scriptsize $\pm$1.3}}\\

\bottomrule
\end{tabular}
\caption{
Closed-set accuracy (Acc.) and overall OOD detection AUC across all experimental configurations.
Results are reported as mean $\pm$ standard deviation over three runs with a fixed random seed.
The best performance is shown in bold, and the second best is underlined.}

\label{tab:performance}
\end{table*}

\subsection{Objectives}
The objective functions used to jointly train the closed-set classifier, the OOD detector, and the SNA module are described below.

\noindent\textbf{Closed-set classifier.}
We adopt the FixMatch~\cite{sohn2020fixmatch} framework to train the closed-set classifier using labeled and unlabeled data. The loss consists of a supervised cross-entropy loss $\mathcal{L}_x$ on labeled samples and a consistency regularization loss $\mathcal{L}_u$ based on unlabeled samples pseudo-labeled via softmax predictions:
\begin{equation}
\mathcal{L}_{\text{CC}} = \mathcal{L}_x(\mathit{X}) + \lambda_{u} \mathcal{L}_u(\mathcal{U})
\end{equation}
The balancing weight $\lambda_u$ controls the relative strength of the unsupervised term.

\noindent\textbf{OOD detector.} 
We follow the loss formulation of OpenMatch~\cite{saito2021openmatch} and SSB~\cite{fan2023ssb} to train the OOD detector. The total loss includes the One-vs-All (OVA) loss $\mathcal{L}_{\text{od}}^{\text{OVA}}$ for labeled data, an entropy minimization loss $\mathcal{L}_{\text{od}}^{\text{em}}$ to sharpen predictions on unlabeled samples, and a soft open-set consistency regularization loss $\mathcal{L}_{\text{od}}^{\text{SOCR}}$ that promotes prediction consistency under augmentation.
In addition, we incorporate a pseudo-negative loss $\mathcal{L}_{\text{od}}^{\text{neg}}$, which leverages confidently identified OOD samples from $\mathcal{U}$ to enhance ID-OOD separation. To further improve robustness, both weak and strong augmentations are applied to these negative samples.
The final OOD detector loss is defined as:
\begin{align}
\mathcal{L}_{\text{OD}} =\,
&\mathcal{L}_{\text{od}}^{\text{OVA}}(\mathit{X})
+ \lambda_{\text{em}} \mathcal{L}_{\text{od}}^{\text{em}}(\mathcal{U}) \notag \\
&+ \lambda_{\text{SOCR}} \mathcal{L}_{\text{od}}^{\text{SOCR}}(\mathcal{U})
+ \lambda_{\text{neg}} \mathcal{L}_{\text{od}}^{\text{neg}}(\mathcal{U})
\end{align}
Each $\lambda$ weights the corresponding loss component.

\begin{figure*}[t]
  \centering
    \includegraphics[width=\linewidth]{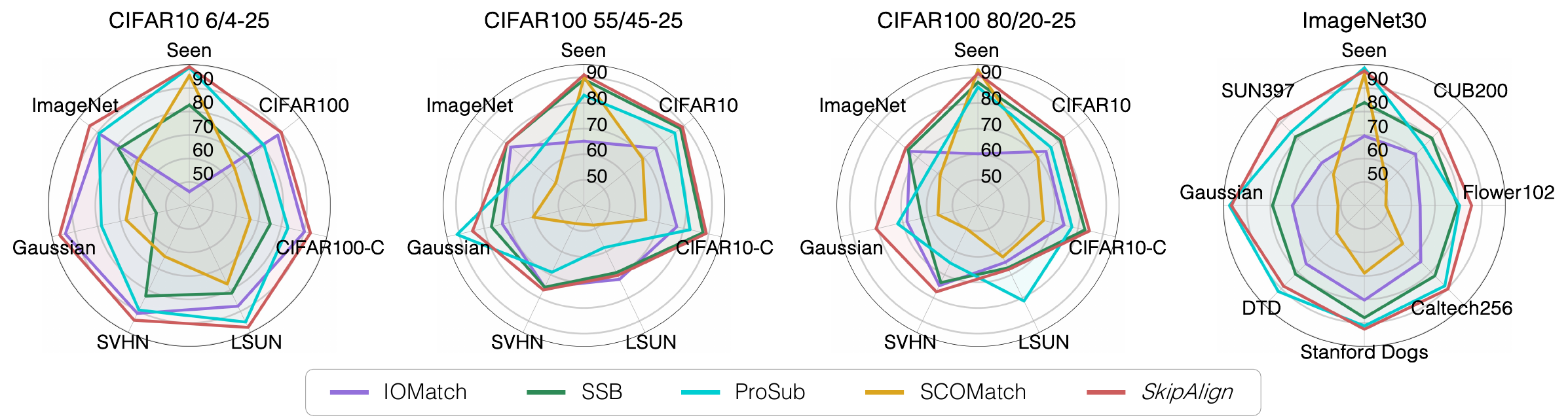}
    \caption{
    Per-dataset OOD detection AUC for various OSSL methods across different experimental configurations
    (CIFAR-10, CIFAR-100 with varying ID/OOD splits and 25 labels, and ImageNet30).
    \proposal consistently achieves strong and balanced detection performance on most OOD datasets.}
    
  \label{fig:auroc}
\end{figure*}

\begin{figure}[ht!]
  \centering
    \includegraphics[width=\linewidth]{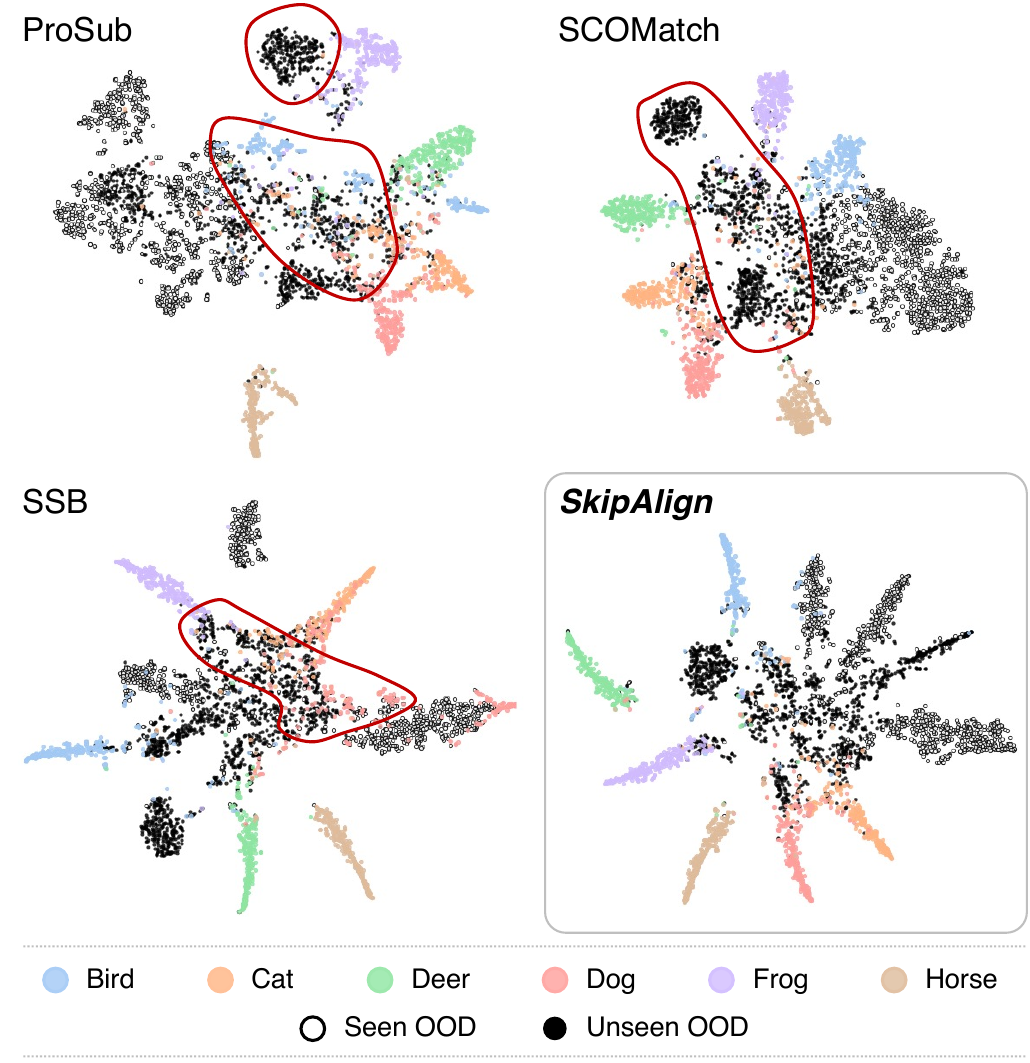}
    \caption{
    t-SNE visualization of feature embeddings from the OOD detector (CIFAR-10, 6/4-50).
    Red contours highlight problematic regions in prior methods: 
    \proposal yields compact, class-specific ID clusters and clear ID–OOD separation, with unseen OODs naturally positioned in non-ID regions or around the OOD boundary.
    }
  \label{fig:features}
\end{figure}


\noindent\textbf{Overall Objective.} 
The final training objective combines the three loss terms:
\begin{equation}
\mathcal{L}_{\text{Total}} = \lambda_{\text{CC}} \mathcal{L}_{\text{CC}}  + \lambda_{\text{OD}} \mathcal{L}_{\text{OD}} + \lambda_{\text{SNA}} \mathcal{L}_{\text{SNA}}
\end{equation}
Here, $\mathcal{L}_{\text{SNA}}$ refers to the selective alignment loss described above, and $\lambda_{\text{cc}}, \lambda_{\text{od}}, \lambda_{\text{SNA}}$ control the relative weights of each objective.
\section{Experiments}
\label{sec:experiments}

\subsection{Experimental Setup}

We evaluate \proposal under standard OSSL protocols, measuring both ID (closed-set) classification accuracy and OOD detection performance using AUC (Area Under the Receiver Operating Characteristic curve). To rigorously assess OOD generalization, we include \textbf{a diverse set of unseen OOD datasets} in addition to the seen OODs used during training.
For overall detection performance, we report \textit{overall AUC}, defined as the mean AUC across all OOD datasets (seen and unseen), weighting each dataset equally.

\noindent\textbf{Datasets and Backbones.}  
For CIFAR-10 and CIFAR-100~\cite{krizhevsky2009learning}, we use WideResNet-28-2~\cite{zagoruyko2016wide} and vary the number of ID/OOD classes with 25 or 50 labeled samples per class. 
For CIFAR-10, six animal classes are selected as ID classes, and the remaining four are used as OOD.  
In CIFAR-100, we first group classes by their semantic super-classes and then assign them into disjoint ID and OOD subsets.  
We evaluate two settings with different OOD proportions: one with 55 ID and 45 OOD classes, and another with 80 ID and 20 OOD classes.  
For each configuration, 25 or 50 labeled samples per ID class are used for training.  
For ImageNet30~\cite{hendrycks2016baseline}, a 30-class subset of ImageNet~\cite{deng2009imagenet}, we adopt ResNet-18~\cite{he2016deep} and use 20 classes as ID and 10 as OOD, with 5\% of the ID data labeled (65 samples per class). 
For TinyImageNet~\cite{le2015tiny}, we use WideResNet-28-4 on a 100 ID / 100 OOD split, with 10\% labeled data per ID class.

\noindent\textbf{Unseen OOD Evaluation.}  
To evaluate the generalization ability to unseen OOD data, we augment the standard seen OOD test with additional unseen OOD sources.
For CIFAR-10 and CIFAR-100, we use CIFAR-100 and CIFAR-10, respectively, as unseen OOD datasets, along with CIFAR-100C and CIFAR-10C~\cite{hendrycks2019robustness}, ImageNet~\cite{deng2009imagenet}, LSUN~\cite{yu2015lsun}, SVHN~\cite{netzer2011reading}, and Gaussian noise. 
For ImageNet30, we include a diverse set of natural and synthetic categories such as Gaussian noise, Stanford Dogs~\cite{KhoslaYaoJayadevaprakashFeiFei_FGVC2011}, CUB-200~\cite{welinder2010caltech}, Flowers102~\cite{Nilsback08}, Caltech-256~\cite{griffin2007caltech}, Describable Textures Dataset (DTD)~\cite{cimpoi2014describing}, LSUN, and SUN397~\cite{xiao2010sun}.  
For TinyImageNet (TIN), we adopt unseen OOD datasets including Gaussian noise, SVHN, DTD, LSUN, and SUN397.

\noindent\textbf{Baselines.} 
We compare \proposal against recent OSSL methods, including IOMatch, SSB, ProSub, and SCOMatch.  
All methods are evaluated under the same dataset splits and hardware environment (H100 or RTX 3090), using official or faithfully reproduced implementations.


\subsection{Main Results}

\noindent\textbf{Performances.} 
Table~\ref{tab:performance} reports closed-set accuracy and overall AUC across all experimental configurations.
\proposal consistently achieves the highest AUC in every configuration, with improvements of \textbf{up to 7.1 points} over the strongest baseline.
Closed-set accuracy is also the best in nearly all settings, with only marginal drops on CIFAR-10 and TinyImageNet, and shows a maximum gain of 2.7 percentage points (pp) compared to competing methods.

Figure~\ref{fig:auroc} further breaks down AUC performance for each OOD dataset (including both seen and unseen sources).
\proposal demonstrates \textbf{consistently strong and balanced detection performance} across nearly all OOD datasets.
One exception is LSUN under the CIFAR-100 setting: as the number of ID classes grows, many scene-centric categories overlap semantically with LSUN, causing feature-level similarity and reduced separability of OOD samples.
Interestingly, ProSub appears to perform well on LSUN in the 80-class CIFAR-100 setting, but this stems from poor ID classification---its closed-set accuracy is 12 pp lower than that of \proposal, indicating that it misclassifies many ID samples rather than truly improving OOD discrimination.
Finally, unlike other methods, whose AUC varies substantially between 25- and 50-label settings, \proposal maintains consistently high OOD detection performance regardless of the number of labeled samples.


\noindent\textbf{t-SNE Analysis of OOD Detector Embeddings.} 
To better understand \proposal's performance gains, 
we visualize and compare how different methods structure embeddings within the OOD detector using t-SNE (Figure~\ref{fig:features}).
ProSub separates seen OODs from ID clusters but leaves unseen OODs nearby, which can be located within the ID subspace and cause mis-detection. Additionally, the cluster for the ID ``Bird'' class remains fragmented due to early misalignment when relying solely on feature-subspace similarity.
SCOMatch defines a synthetic OOD class that appears as a distinct region in the embedding space, yet most unseen OODs fall outside this predefined region and are misclassified as ID. 
SSB produces mostly compact ID clusters, but the ``Dog'' class lacks clear boundaries and overlaps with nearby seen or unseen OODs. Unseen OODs tend to gather near the center and intrude into ID regions, reflecting the limitations of OVA classifiers trained with very limited labeled data. 

In contrast, \proposal leverages SNA to better structure the embedding space.
The SNA module rotates ID samples toward their corresponding class prototypes, facilitating class alignment, while uncertain or OOD samples are rotated toward directions orthogonal to all prototypes, preventing OOD data from being assigned to any class.
Through this mechanism, ID sample embeddings naturally condense around their class prototypes, while OOD samples reside in the voids outside the regions occupied by ID classes.


\subsection{Ablation} 

\noindent
\textbf{Selective Non-Alignment Loss Variants.}
We assess the effectiveness of our Unlabeled SNA loss ($\mathcal{L}_{\text{USNA}}$, Eq.~\ref{eqn:eq2}) by comparing it with two alternatives in Table~\ref{tab:loss_comparison}.
The first variant, \textit{OVA with Confident ID}, augments OVA classifier training by treating confidently filtered ID samples from unlabeled data as additional positives. 
This approach significantly lowers both accuracy and AUC compared to ours, suggesting that even small amounts of label noise can destabilize the one-vs-all decision boundaries that are designed to rely on clean labeled data.
The second variant, \textit{Cosine Similarity}, only pulls confident samples toward prototypes without repelling uncertain ones (i.e., excluding them from training). On the 55/45-25 setting, its accuracy is close to ours, but on the 80/20-25 setting it drops by about 1.3 pp, and AUC is consistently lower in both settings. This indicates that selective attraction alone (pull only) cannot maintain sufficient separation between ID and OOD embeddings.
By contrast, $\mathcal{L}_{\text{USNA}}$ achieves the best balance across both settings, improving both accuracy and OOD detection performance. Its selective attraction combined with mild repulsion of uncertain features yields a more structured and discriminative feature space for robust OOD detection.

\begin{table}[t]
\centering
\begingroup
\setlength{\tabcolsep}{3.5mm}   
\small 

\begin{tabular}{ccccc}
\toprule
\multirow{2}{*}{\textbf{Loss}} & 
\multicolumn{2}{c}{\textbf{80/20-25}} &
\multicolumn{2}{c}{\textbf{55/45-25}} \\
\cmidrule(lr){2-3}\cmidrule(lr){4-5}
 & \textbf{Acc.} & \textbf{AUC} & \textbf{Acc.} & \textbf{AUC} \\
\midrule
OVA w/ Conf. ID                    & 64.4 & 69.7 & 68.5 & 69.7 \\
Cosine Similarity                  & 65.9 & 79.4 & 71.4 & 83.4 \\
$\mathcal{L}_{\text{USNA}}$ (Ours) & \textbf{67.2} & \textbf{80.8} & \textbf{71.8} & \textbf{84.1} \\
\bottomrule
\end{tabular}
\caption{Comparison of Unlabeled SNA loss variants on CIFAR-100 under two dataset configurations.}
\label{tab:loss_comparison}
\endgroup
\end{table}



\noindent
\textbf{SNA Loss Combination.}
\begin{table}[t]
\centering
\begingroup
\setlength{\tabcolsep}{1.8mm} 
\small

\renewcommand{\arraystretch}{1.1}
\begin{tabular}{c ccc c ccc}
\toprule
\multirow{2}{*}{\textbf{Exp.}} &
\multicolumn{3}{c}{\textbf{Loss Combination}} &
\multirow{2}{*}{\textbf{Acc.}} &
\multicolumn{3}{c}{\textbf{AUC}} \\
\cmidrule(lr){2-4}\cmidrule(lr){6-8}
 & IA & PA & USNA &  & Seen & Unseen & Overall \\
\midrule
1 &            &            &            & 64.7 & 88.6 & 76.2 & 78.0 \\
2 & \checkmark & \checkmark &            & 67.2 & \textbf{91.5} & 77.7 & 79.7 \\
3 &            &            & \checkmark & 67.2 & 89.9 & 78.3 & 80.0 \\
4 & \checkmark & \checkmark & \checkmark & \textbf{67.2} & 91.2 & \textbf{79.1} & \textbf{80.9} \\
\bottomrule
\end{tabular}

\caption{Ablation on SNA loss combinations on CIFAR-100 80/20-25.
}
\label{tab:loss_combination}
\endgroup
\end{table}
Table~\ref{tab:loss_combination} evaluates the contribution of each loss component within the SNA module: 
$\mathcal{L}_{\text{IA}}$, $\mathcal{L}_{\text{PA}}$, and $\mathcal{L}_{\text{USNA}}$. 
Without these losses (Exp.~1), both accuracy and AUC are low due to poor feature structuring.
Supervised alignment alone (Exp.~2) improves accuracy and seen OOD detection but fails to leverage unlabeled data, limiting the generalization ability of OOD detection.
USNA alone (Exp.~3) slightly improves overall AUC over Exp.~2 by avoiding false alignment of uncertain samples, but lacking supervised alignment degrades ID cluster quality and lowers seen AUC.
Combining all three losses (Exp.~4) yields the best performance: labeled losses form compact and well-separated ID clusters, while USNA keeps uncertain samples away from prototypes, enlarging ID–OOD margins. 
This synergy achieves the highest total AUC, highlighting the benefit of combining supervised alignment with selective non-alignment for robust OOD detection.

\noindent
\noindent\textbf{Prototype Refinement.}
\begin{figure}[t]
  \centering
    \includegraphics[width=0.99\linewidth]{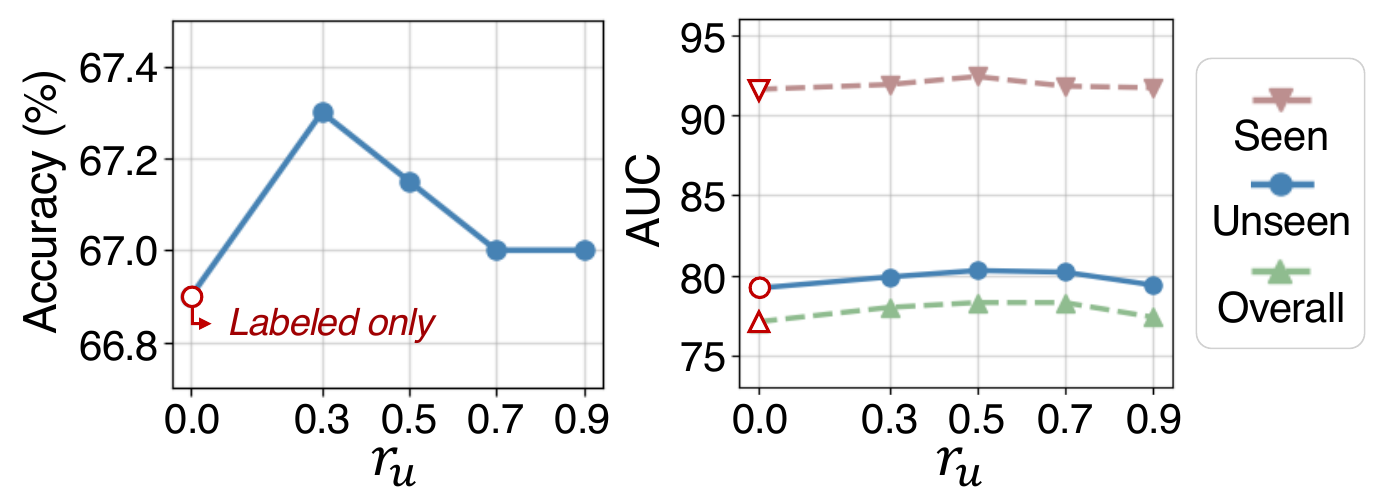}
    \caption{Effect of unlabeled contribution $r_u$ in prototype refinement on CIFAR-100 80/20-25.}
\label{fig:ulbcontribution}
\end{figure}
%
We analyze the effect of the unlabeled contribution ratio $r_u$, a weighting parameter in our adaptive prototype generation that determines how much confidently pseudo-labeled samples contribute relative to labeled data.
Figure~\ref{fig:ulbcontribution} shows that increasing $r_u$ from 0 (labeled only) to a moderate value (around 0.3–0.5) improves both accuracy and overall OOD AUC, indicating that leveraging reliable unlabeled samples helps refine prototypes and leads to a more stable decision geometry that generalizes beyond seen OODs.
However, setting $r_u$ too high introduces residual pseudo-label noise, causing slight performance degradation.

\noindent
\textbf{ID Selection Threshold.}  
\begin{table}[t]
\centering
\begingroup
\setlength{\tabcolsep}{3.5mm}   
\small 

\begin{tabular}{ccccc}
\toprule
\multirow{2}{*}{\textbf{$\eta_{\scriptscriptstyle ID}$}} & 
\multicolumn{2}{c}{\textbf{CIFAR10 6/4-25}} &
\multicolumn{2}{c}{\textbf{CIFAR100 55/45-25}} \\
\cmidrule(lr){2-3}\cmidrule(lr){4-5}
 & \textbf{Acc.} & \textbf{AUC} & \textbf{Acc.} & \textbf{AUC} \\
\midrule
0.0 & 91.4 & 80.0 & 70.9 & 82.5 \\
0.3 & 91.6 & 89.5 & 71.5 & 83.1 \\
0.5 & 92.1 & \textbf{94.4} & \textbf{71.8} & 83.2 \\
0.7 & \textbf{92.4} & 91.4 & 71.5 & \textbf{85.1} \\
0.9 & 91.5 & 89.8 & 70.8 & 83.1 \\
\bottomrule
\end{tabular}

\caption{Effect of the OOD detector threshold $\eta_{ID}$ in dual-gate ID selection 
($\tau_{ID}=0.99$, temperature $0.5$).}
\label{tab:od_threshold}
\endgroup
\end{table}
We analyze the impact of varying the OOD confidence threshold $\eta_{\text{ID}}$ in our dual-gate ID selection, with the classifier threshold fixed at $\tau_{\text{ID}} = 0.99$ and softmax temperature 0.5 (Table~\ref{tab:od_threshold}).
All nonzero thresholds outperform the classifier-only gating ($\eta_{\text{ID}} = 0$), demonstrating the importance of incorporating the OOD detector for reliable ID selection.
On both CIFAR-10 and CIFAR-100, performance steadily improves as $\eta_{\text{ID}}$ increases from 0 and remains stable around moderate values (e.g., 0.5–0.7).
While overly conservative settings (e.g., $\eta_{\text{ID}} = 0.9$) may slightly reduce accuracy by filtering out some useful ID samples,
this behavior reflects the binary nature of ID/OOD separation and indicates that our method performs robustly across a wide range of threshold values.

\section{Conclusion}

We introduced \proposal, an OSSL framework based on Selective Non-Alignment (SNA) to improve generalization beyond seen OOD samples. SNA avoids overfitting by skipping alignment for uncertain samples and applying angular repulsion, encouraging OOD features to remain in a separate void. Combined with dual-gate filtering and prototype refinement, \proposal effectively learns under label noise. Experiments show strong AUC and generalization gains over prior methods while maintaining competitive accuracy. Although performance is robust across diverse OODs, challenges remain when unseen classes closely resemble ID ones (e.g., LSUN under the CIFAR-100 setting). Future work includes enhancing feature disentanglement for such cases.

\section{Acknowledgments}
This work was supported by the Information and Communications Technology Planning and Evaluation (IITP) grant funded by the Korea Ministry of Science and ICT (MSIT) (Grant No. 2710085245); the National Research Foundation of Korea (NRF) grant funded by the Korea government (MSIT) through the Engineering Research Center (ERC) program (Grant No. 2710077975); and by the Mobile eXperience (MX) Business, Samsung Electronics Co., Ltd. Hyung-Sin Kim is the corresponding author.

\bibliography{main}

\clearpage
\setcounter{page}{1}

\renewcommand{\thesection}{\Alph{section}} 
\setcounter{section}{1}

\twocolumn[
  \begin{center}
    \LARGE\bfseries Let the Void Be Void: Robust Open-Set Semi-Supervised Learning via\\Selective Non-Alignment \\ \normalfont Supplementary Material
  \end{center}
]


\section{A. Analysis of Feature Dynamics in SNA}
To support our claim that \textbf{Selective Non-Alignment (SNA)} promotes robust feature structuring by 
\emph{deliberately redirecting the feature dynamics}
for uncertain samples, we conduct a gradient-based analysis of the feature dynamics.  
Unlike conventional contrastive methods that align all samples toward class prototypes, SNA explicitly \emph{prevents alignment} for uncertain and OOD samples ($\Phi_i = 0$), applying only angular repulsion without directional attraction.  
This selective non-alignment complements the classifier ($\mathcal{L}_x$, $\mathcal{L}_u$) and OVA detector ($\mathcal{L}_{\text{od}}^{\text{OVA}}$), mitigating their tendency to amplify norm even for noisy or misclassified features.

\subsection{A.1 Setup and Notation}

Let $\mathcal{F}$ be the shared backbone network producing a feature embedding $f \in \mathbb{R}^{d_f}$. This embedding is processed by:
\begin{itemize}
    \item \textbf{Closed-set classifier} $\mathcal{H}_{\text{cls}}$: logits $h_{\text{cls}} = \mathcal{H}_{\text{cls}}(f)$.
    \item \textbf{OOD detector} $\mathcal{H}_{\text{ova}}$: $K$ one-vs-all binary logits $h_{\text{ova}} = \mathcal{H}_{\text{ova}}(f)$.
    \item \textbf{SNA module} $\mathcal{P}$: projected vector $z = \mathcal{P}(f) \in \mathbb{R}^{d}$ for SNA.
\end{itemize}

We analyze the total gradient flow on $f$ under the loss:
\[
\mathcal{L}_{\text{Total}} = \lambda_{\text{CC}} \mathcal{L}_{\text{CC}}  + \lambda_{\text{OD}} \mathcal{L}_{\text{OD}} + \lambda_{\text{SNA}} \mathcal{L}_{\text{SNA}}
\]

\subsection{A.2 Closed-set Classifier Loss}

The classifier head $\mathcal{H}_{\text{cls}}$ receives the shared feature $f \in \mathbb{R}^{d_f}$ and computes class logits via an MLP. We analyze the gradients induced by its two loss components: the cross-entropy loss $\mathcal{L}_x$ for labeled samples and the consistency loss $\mathcal{L}_u$ for unlabeled samples.

\paragraph{Cross-Entropy Loss Gradient.}
Let $w_j \in \mathbb{R}^{d_f}$ denote the classifier weight for class $j$. The logit is computed as:
\[
\ell_j = f^\top w_j,
\]
and the softmax probability is given by:
\[
\alpha_j = \frac{\exp(\ell_j)}{\sum_{m=1}^K \exp(\ell_m)}.
\]
Given a ground-truth label $k$, the cross-entropy loss becomes:
\[
\mathcal{L}_x = -\log \alpha_k,
\]
with gradient:
\[
\nabla_f \mathcal{L}_x = \sum_{j=1}^K (\alpha_j - \delta_{jk}) w_j,
\]
where $\delta_{jk}$ is the Kronecker delta.

To analyze how this gradient affects feature norm, we project it onto the unit vector $\hat{f} = f / \|f\|$:
\[
\frac{d}{dt} \|f\| = -\hat{f}^\top \nabla_f \mathcal{L}_x = -\sum_{j=1}^K (\alpha_j - \delta_{jk}) \|w_j\| \cos \theta_j,
\]
where $\theta_j$ is the angle between $f$ and $w_j$. This term is typically positive (i.e., leads to norm increase) when the prediction is confident ($\alpha_k \approx 1$) and aligned with $w_k$ (i.e., $\cos \theta_k > 0$).

\paragraph{Consistency Loss Gradient.}
For an unlabeled sample, let $\tilde{\alpha} \in \mathbb{R}^K$ be the fixed pseudo-label (e.g., from weak augmentation), and $\alpha$ be the predicted softmax probability (e.g., from strong augmentation). The consistency loss is:
\[
\mathcal{L}_u = -\sum_{j=1}^K \tilde{\alpha}_j \log \alpha_j,
\]
with gradient:
\[
\nabla_f \mathcal{L}_u = \sum_{j=1}^K (\alpha_j - \tilde{\alpha}_j) w_j.
\]
This gradient also exhibits radial structure toward classifier weights and generally increases $\|f\|$. However, since $\tilde{\alpha}$ may be noisy or incorrect, the resulting pull may amplify norm in misleading directions.

\noindent\textbf{Remark.} Both $\mathcal{L}_x$ and $\mathcal{L}_u$ produce radial gradients directed toward classifier weights. When predictions are confident, these gradients increase feature norm. Without additional regularization, even mislabeled or ambiguous samples can experience norm growth, motivating the need for SNA to regulate this effect.

\subsection{A.3 OOD Detector Loss}

We focus on the main OOD detection loss $\mathcal{L}_{\text{od}}^{\text{OVA}}$, which is based on a one-vs-all (OVA) classification scheme.  
Each of the $K$ binary sub-classifiers predicts whether a sample belongs to class $k$ (ID) or not (OOD) by outputting a probability pair $(\varphi_k^{ID}, \varphi_k^{OOD})$ from the corresponding logits $(s_k^{ID}, s_k^{OOD})$:
\[
(s_k^{ID}, s_k^{OOD}) = W_k f + b_k,
\]
where $W_k \in \mathbb{R}^{2 \times d_f}$ and $b_k \in \mathbb{R}^2$ are parameters for class $k$.

\paragraph{OVA Loss Gradient.}
Given a labeled sample $(x_i, y_i)$, the binary cross-entropy loss for class $k$ is:
\[
\mathcal{L}_{\text{ova}}^{(i, k)} =
- \mathds{1}(y_i = k) \log \sigma(s_k^{\text{ID}})
- \mathds{1}(y_i \neq k) \log \sigma(s_k^{\text{OOD}}),
\]
where $\sigma(\cdot)$ is the sigmoid function.  
The full OVA loss is:
\[
\mathcal{L}_{\text{od}}^{\text{OVA}} = \sum_i \sum_{k=1}^K \mathcal{L}_{\text{ova}}^{(i, k)}.
\]
The gradient with respect to $f$ is:
\[
\nabla_f \mathcal{L}_{\text{ova}}^{(i, k)} =
\begin{cases}
-(1 - \sigma(s_k^{\text{ID}})) \cdot w_k^{\text{ID}}, & \text{if } y_i = k \\[0.5em]
- \sigma(s_k^{\text{OOD}}) \cdot w_k^{\text{OOD}}, & \text{if } y_i \neq k
\end{cases}
\]
where $w_k^{\text{ID}}$, $w_k^{\text{OOD}}$ are the corresponding row vectors of $W_k$.  
The impact on the feature norm is given by:
\[
\frac{d}{dt} \|f\| = -\hat{f}^\top \nabla_f \mathcal{L}_{\text{ova}}^{(i, k)},
\]
which typically increases $\|f\|$ when the prediction is confident. This promotes separation between relevant and irrelevant class directions, reinforcing inter-class margins.

\subsection{A.4 SNA Loss: Angular Update}

Among the components of the SNA loss, we focus on the $\mathcal{L}_{\text{USNA}}$, which plays a key role in regulating norm growth for uncertain samples.  
This loss is defined over the projected embedding $z = \mathcal{P}(f)$, and its gradient backpropagates to the backbone feature $f$ through the projection head $\mathcal{P}(\cdot)$.

\paragraph{USNA Loss Gradient.}
Let $J_{\mathcal{P}} = \frac{\partial z}{\partial f} \in \mathbb{R}^{d \times d_f}$ denote the Jacobian matrix of the projection MLP. Then, the gradient with respect to $f$ is given by:
\[
\nabla_f \mathcal{L}_{\text{USNA}} = J_{\mathcal{P}}^\top \nabla_z \mathcal{L}_{\text{USNA}}.
\]
The gradient with respect to the normalized projection $\hat{z}_i = z_i / \|z_i\|$ is given in Eq.~\ref{eq:grad_z} of the main paper, and is repeated below for convenience:
\[
\nabla_{z_i} \mathcal{L}_{\text{USNA}} = \frac{1}{T \|z_i\|} \left( I - \hat{z}_i \hat{z}_i^\top \right) \left( \sum_{j=1}^K \alpha_j \hat{\mu}_j - \Phi_i \hat{\mu}_k \right).
\]
Here, $\hat{\mu}_j$ is the normalized prototype of class $j$, $\alpha_j$ is the soft pseudo-label probability for class $j$, and $\Phi_i \in \{0, 1\}$ is the dual-gate confidence mask.
This gradient is orthogonal to $\hat{z}_i$, since:
\[
z_i^\top \nabla_{z_i} \mathcal{L}_{\text{USNA}} = 0,
\]
which implies it is \textbf{purely angular}. 

\paragraph{Effect on Backbone Feature $f$.}
Because the projection MLP $\mathcal{P}$ is smooth and approximately preserves direction in early training, the tangential property of $\nabla_{z_i} \mathcal{L}_{\text{USNA}}$ is partially retained after backpropagation to $f$.  
That is, $\nabla_f \mathcal{L}_{\text{USNA}}$ primarily induces a rotational (re-orienting) force rather than radial expansion, leading to directional adjustment 
that steers the trajectory of
$\|f\|$.
\subsection{A.5 Combined Dynamics and the Role of SNA}

We now synthesize the effects of all loss components---the cross-entropy loss $\mathcal{L}_x$, the consistency loss $\mathcal{L}_u$, the OVA loss $\mathcal{L}_{\text{od}}^{\text{OVA}}$, and the SNA loss $\mathcal{L}_{\text{USNA}}$---under the dual-gate mask $\Phi_i \in \{0, 1\}$ that determines whether a sample is confidently identified as ID.

\paragraph{Confident Samples ($\Phi_i = 1$).}
For confidently predicted ID samples, all four losses act in a consistent direction:
\begin{itemize}[leftmargin=*, noitemsep, topsep=0pt]
    \item $\mathcal{L}_x$ pulls the feature $f$ toward the classifier weight $w_k$.
    \item The consistency loss $\mathcal{L}_u$ applies a similar force based on a confident pseudo-label, reinforcing the classifier gradient.
    \item The OVA loss $\mathcal{L}_{\text{od}}^{\text{OVA}}$ pulls $f$ toward the OVA embedding $w_k^{\text{ID}}$.
    \item The SNA loss $\mathcal{L}_{\text{USNA}}$ aligns the projection $z = \mathcal{P}(f)$ toward the class prototype $\hat{\mu}_k$ via an angular pull.
\end{itemize}

While $\mathcal{L}_{\text{USNA}}$ exerts a \textit{purely angular} force in the projected space, its alignment effect enhances the radial gradients from $\mathcal{L}_x$, $\mathcal{L}_u$, and $\mathcal{L}_{\text{od}}^{\text{OVA}}$, which directly amplify $\|f\|$.
This synergy leads to coherent norm growth and compact, high-norm clusters in both $f$ and $z$ spaces.
Importantly, although the pull directions differ in parameter space---with $\mathcal{L}_x$ and $\mathcal{L}_{\text{od}}^{\text{OVA}}$ pulling toward classifier weights and $\mathcal{L}_{\text{USNA}}$ pulling toward prototypes---these directions converge over training.

Let $\mu_k = \mathbb{E}_{f \sim p_k} [f]$ be the mean feature embedding for class $k$, and let $w_k$ and $w_k^{\text{ID}}$ denote the weight vectors in the classifier and OVA head respectively.
The cross-entropy loss $\mathcal{L}_x$ and OVA loss $\mathcal{L}_{\text{od}}^{\text{OVA}}$ encourage:
\[
\mathbb{E}_{f \sim p_k} [f] \approx \lambda_w w_k \approx \lambda_{\text{OVA}} w_k^{\text{ID}} \quad \text{for some } \lambda_w, \lambda_{\text{OVA}} > 0,
\]
while the SNA loss encourages:
\[
\hat{z}_i = \frac{\mathcal{P}(f)}{\|\mathcal{P}(f)\|} \to \hat{\mu}_k = \frac{\mu_k}{\|\mu_k\|}.
\]
Thus, $w_k$, $w_k^{\text{ID}}$, and $\hat{\mu}_k$ all converge toward the same semantic direction defined by the class-conditional mean $\mu_k$.

As a result, the angular alignment induced by $\mathcal{L}_{\text{USNA}}$ reduces the angle between $z$ and $\hat{\mu}_k$, indirectly steering $f$ toward a direction more aligned with $w_k$ and $w_k^{\text{ID}}$.
This alignment increases $\cos \theta_k$, thereby amplifying the norm growth induced by the Closed-set Classifier and OOD Detector losses.

\paragraph{Uncertain or OOD Samples ($\Phi_i = 0$).}
In contrast, for samples with low ID confidence (i.e., uncertain or OOD), the loss components exert competing gradients that may destabilize the embedding:
\begin{itemize}[leftmargin=*, noitemsep, topsep=0pt]
    \item The consistency loss $\mathcal{L}_u$ is applied to unlabeled samples that pass a softmax-confidence threshold, without explicit OOD filtering. As a result, uncertain or OOD features may be pulled toward incorrect classifier weights $w_k$ based on noisy or ambiguous pseudo-labels, leading to \textit{undesired norm growth}.
    
    \item In contrast, the USNA loss $\mathcal{L}_{\text{USNA}}$ removes all attractive pull and instead applies a \textit{non-selective angular repulsion} from all class prototypes, which preserves the projection norm $\|z_i\|$.
    
    \item This purely angular update in the projected space $z_i = \mathcal{P}(f_i)$ is backpropagated through the projection Jacobian $J_{\mathcal{P}} = \frac{\partial z}{\partial f}$, inducing a tangential force on the backbone feature $f$ 
    that redirects this norm growth.
    
    This component acts as a corrective signal that can \textit{partially cancel the radial pull} from the consistency loss $\mathcal{L}_u$.

    In practice, $\mathcal{P}(\cdot)$ is implemented as a non-linear MLP (e.g., with ReLU), which introduces a data-dependent Jacobian. This allows the angular signal $\nabla_z \mathcal{L}_{\text{USNA}}$ to be mapped into a broader set of directions in the $f$-space while still avoiding radial amplification.

    Concretely, if we assume a locally linear projection head $\mathcal{P}(f) \approx A f$, the SNA gradient becomes:
    \[
    \nabla_f \mathcal{L}_{\text{USNA}} \approx A^\top \nabla_z \mathcal{L}_{\text{USNA}}, \quad \text{with} \quad \nabla_z \mathcal{L}_{\text{USNA}} \perp \hat{z}.
    \]
    Since $\hat{z} \approx A \hat{f}$, it follows that:
    \[
    \hat{f}^\top \nabla_f \mathcal{L}_{\text{USNA}} \approx 0,
    \]
    implying that the SNA gradient has minimal radial projection in the $f$ space.
    
    \item Furthermore, the repulsion term $\sum_j \alpha_j \hat{\mu}_j$ in the SNA loss applies a uniform angular push away from all class prototypes. This prevents $z_i$ from aligning with any specific prototype, maintaining a low $\cos \theta_k$ for all $k$. Since Closed-set Classifier and OOD Detector losses amplify norm in proportion to $\cos \theta_k$, 
    
    this misalignment redirects their norm-growing effect away from prototypes:
    \[
    \frac{d}{dt} \|f\| \propto \cos \theta_k \cdot \|\nabla_f \mathcal{L}_x\| \quad \text{(similarly for $\mathcal{L}_u$, $\mathcal{L}_{\text{od}}^{\text{OVA}}$)}.
    \]
\end{itemize}

As a result, norm growth for uncertain or OOD features 
is applied in a repulsive direction
, and such embeddings remain in a relatively low-norm, diffuse region. This mechanism is essential for maintaining a clear separation between ID and OOD samples, as illustrated in Figure~\ref{fig:sna}.
\paragraph{SNA Structures \boldmath{$f$} through Angular Supervision on \boldmath{$z$}.}
Although the USNA loss operates solely in the projected space $z = \mathcal{P}(f)$, its gradients influence the backbone feature $f$ via backpropagation through the projection head. As training progresses:
\begin{itemize}[leftmargin=*, noitemsep, topsep=0pt]
    \item Confident ID samples are pulled toward class prototypes, becoming aligned in both direction and norm.
    \item Uncertain or OOD samples remain misaligned with all prototypes, with their norm suppressed due to tangential SNA updates.
\end{itemize}

\subsection{A.6 Theoretical Analysis of Generalization}
In this section, we theoretically analyze the generalization properties of our proposed approach versus the conventional (K+1)-way classification method for OOD detection. To formalize this comparison, we utilize Rademacher complexity.

\noindent\textbf{Theorem 1.} \emph{Let $\mathcal{H}_{K+1}$ be a hypothesis space of functions $h:\mathcal{X}\rightarrow\mathbb{R}^{K+1}$ for $(K+1)$-way classification, where the $(K+1)$-th class is a synthetic representation for all Out-of-Distribution (OOD) data. Let $\mathcal{L}_{CE}$ be the standard multi-class cross-entropy loss. For any hypothesis $h\in\mathcal{H}_{K+1}$, with probability at least $1-\delta$ over a sample S of size m, the true risk $\mathcal{R}(h)$ is bounded by:}
\begin{equation*}
\mathcal{R}(h)\le\hat{\mathcal{R}}_{S}(h)+2\mathfrak{R}_{m}(\mathcal{L}_{CE}\circ\mathcal{H}_{K+1})+\mathcal{O}(\sqrt{\frac{\log(1/\delta)}{m}})
\end{equation*}
\emph{where $\hat{\mathcal{R}}_{S}(h)$ is the empirical risk and $\mathfrak{R}_{m}$ is the Rademacher complexity.}

The Empirical Risk Minimization (ERM) principle for the $(K+1)$-way model compels the learning algorithm to map the inherently heterogeneous and expansive OOD data distribution, $P_{OOD}$, into a single, compact representation corresponding to the $(K+1)$-th class. This ``aggressive alignment" results in a ``geometric collapse" of the OOD space. Consequently, the hypothesis space $\mathcal{H}_{K+1}$ must possess a high capacity to represent the complex and non-linear decision boundaries required to separate the $K$ In-Distribution (ID) clusters from this artificially constructed OOD cluster. A hypothesis space of high capacity inherently exhibits a large Rademacher complexity, $\mathfrak{R}_{m}(\mathcal{L}_{CE}\circ\mathcal{H}_{K+1})$. This large complexity term yields a loose generalization bound, offering weak guarantees on the model's performance, particularly for unseen OOD samples that do not conform to the collapsed geometry learned from seen OODs.

\noindent\textbf{Theorem 2.} \emph{Let $f_{\theta}:\mathcal{X}\rightarrow\mathcal{Z}$ be a feature extractor trained with the Selective Non-Alignment (SNA) objective, where $\mathcal{Z}\subset\{z\in\mathbb{R}^{d}:||z||\le R_{feat}\}$. Assume the SNA dynamics induce a composite geometric margin $\gamma_{SNA}$ between ID and OOD distributions within $\mathcal{Z}$. Let $\mathcal{G}_{\gamma}$ be the hypothesis space of OOD detectors $g:\mathcal{Z}\rightarrow\mathbb{R}$ that separate ID and OOD data with margin $\gamma_{SNA}$. For any $g\in\mathcal{G}_{\gamma}$, with probability at least $1-\delta$ over a sample S of size m, the true risk $\mathcal{R}(g\circ f_{\theta})$ is bounded by:}
\begin{equation*}
\label{eqn:eq21}
\mathcal{R}(g\circ f_{\theta})\le\hat{\mathcal{R}}_{S,\gamma}(g\circ f_{\theta})+\mathcal{O}(\frac{R_{feat}}{\gamma_{SNA}\sqrt{m}}+\sqrt{\frac{\log(1/\delta)}{m}})
\end{equation*}
\emph{where $\hat{\mathcal{R}}_{S,\gamma}$ is the empirical margin loss.}

\noindent The SNA framework reframes the OSSL problem from direct classification to representation learning. Its objective is to engineer a feature space that maximizes a composite geometric margin, $\gamma_{SNA}$, composed of both norm-based and angle-based separation. The gradient dynamics of the SNA loss, $\mathcal{L}_{SNA}$, mechanistically achieve this.

As discussed in Appendix A.5, confident ID samples receive synergistic alignment forces (from $\mathcal{L}_{CC}$, $\mathcal{L}_{OD}$, and $\mathcal{L}_{SNA}$), leading to rapid norm growth in the correct class direction. Conversely, uncertain or OOD samples receive a selective angular repulsion (from $\mathcal{L}_{USNA}$). As established in Appendix A.4 , this repulsion just  induces slow norm growth, and critically, it does so in a direction \emph{away} from all ID prototypes.

This dynamic—rapid, aligned norm growth for ID vs. slower, repulsive norm growth for OOD—is precisely what maximizes the geometric margin $\gamma_{SNA}$. According to margin-based learning theory, the existence of a large margin $\gamma_{SNA}$ permits the downstream OOD detection task to be solved by a low-capacity hypothesis space $\mathcal{G}_{\gamma}$ (e.g., simple linear classifiers). The Rademacher complexity of such a space is inversely proportional to the margin. Since the SNA algorithm is explicitly designed to maximize $\gamma_{SNA}$, it concurrently minimizes the complexity term of the generalization bound. This results in a significantly tighter bound, ensuring robust generalization to unseen OOD data.

\subsection{A.7 Conclusion}
The SNA module introduces an angular corrective signal that selectively 
redirects the feature dynamics and resulting norm growth for uncertain samples.
While classifier-based losses (e.g., cross-entropy, consistency, and OVA) typically promote norm growth via radial attraction---even for mislabeled or ambiguous samples---USNA applies angular repulsion that primarily steers uncertain features away from ID clusters.
The resulting feature geometry is well-structured: confident ID samples converge into compact, high-norm clusters, while uncertain and OOD samples remain dispersed in low-norm regions. This contrastive separation enhances robustness against pseudo-label noise and strengthens generalization to unseen OODs.


\section{B. Prototype Generation}
As described in ``Prototype Refinement'', we generate adaptive prototypes $\mu_k$ for each class $k$ by combining labeled prototypes $\mu_{lk}$ and unlabeled prototypes $\mu_{uk}$ through a weighted sum. The weights are primarily determined by the number of labeled and unlabeled samples contributing to each prototype. To ensure reliable prototype generation, we assign higher importance to labeled data while adaptively incorporating unlabeled samples based on their trustworthiness.

Let $n_{lk}$ and $n_{uk}$ denote the number of labeled and unlabeled samples used to compute the prototypes for the $k$-th class, respectively. While $n_{lk}$ remains constant, $n_{uk}$ varies dynamically as only trustworthy unlabeled samples, identified through $CC(\cdot)$ and $OD(\cdot)$, are included. The initial weights are calculated as follows:
\vspace{-1ex}
\begin{equation*}
\label{eqn:eq17}
w_{lk} = \gamma \cdot n_{lk}, \quad w_{uk} = r_u \cdot n_{uk}
\end{equation*}
Here, $\gamma$ represents the ratio of labeled to unlabeled batch sizes, and $r_u$ is a hyperparameter controlling the maximum contribution of unlabeled data.
To normalize these weights for proper scaling, we compute:
\vspace{-1ex}
\begin{equation*}
\label{eqn:eq18}
w_{lk}^\text{norm} = \frac{w_{lk}}{w_{lk} + w_{uk}}, \quad w_{uk}^\text{norm} = \frac{w_{uk}}{w_{lk} + w_{uk}}
\end{equation*}
Finally, the adaptive prototype $\mu_k$ is computed as:
\vspace{-1ex}
\begin{equation*}
\label{eqn:eq19}
\mu_k = w_{lk}^\text{norm} \cdot \mu_{lk} + w_{uk}^\text{norm} \cdot \mu_{uk}
\end{equation*}


\section{C. Qualitative Feature Embedding Analysis}

\subsection{C.1 CIFAR-10 6/4-50}
\begin{figure}[t]
  \centering
    \includegraphics[width=\linewidth]{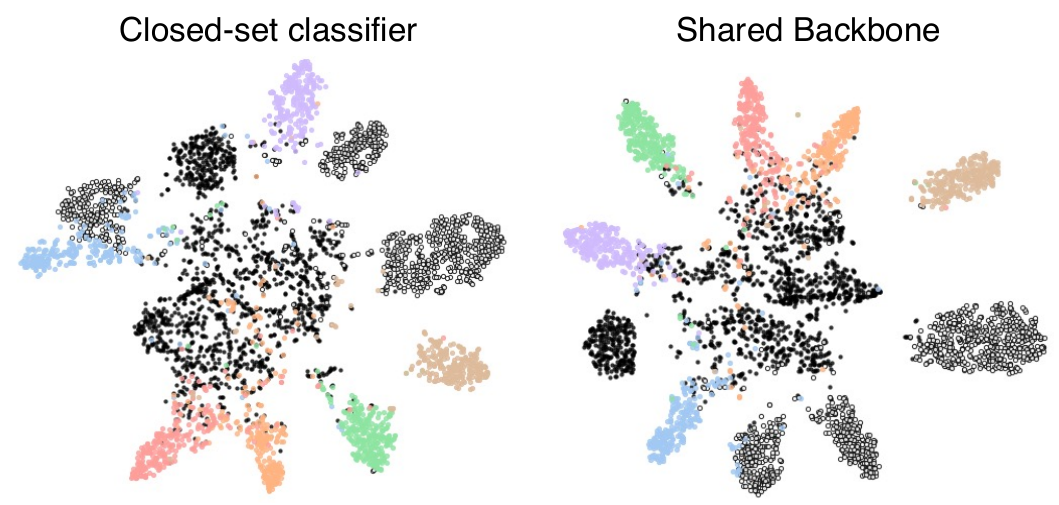}
    \caption{t-SNE visualization of feature embeddings from the Shared Backbone and Closed-sest Classifier (CIFAR10 6/4-50).
    }
  \label{fig:cifar10_features}
\end{figure}

We previously analyzed the feature embeddings of the SNA module and OOD detector on the CIFAR-10 6/4-50 setting. To further examine how different objectives shape the feature space, we visualize the representations from the closed-set classifier and shared backbone in Figure~\ref{fig:cifar10_features}.

The left plot shows the classifier output, trained with cross-entropy for labeled data and consistency loss on pseudo-labeled samples.
While the OOD detector learns independent binary classifiers for each class (OVA), resulting in radially extended, class-specific directions, the closed-set classifier emphasizes relative logit comparisons, encouraging more competitive and mutually exclusive ID clusters. As a result, rounded ID clusters emerge, and seen OOD samples—often misclassified as pseudo-ID—tend to lie near the cluster boundaries. These seen OODs can act as natural augmentations, potentially improving classification performance, but also blur the boundary between ID and OOD.

The shared backbone (right) reflects the combined influence of all objectives—including the classifier, OOD detector, and SNA module. ID clusters remain moderately compact, but OOD samples are more clearly isolated in a diffuse region. This separation arises from the angular repulsion of uncertain samples by SNA and the binary discrimination enforced by the OOD detector.
These losses collectively shape the backbone representation, which in turn feeds into each head, enabling synergistic learning across the entire architecture.

\subsection{C.2 CIFAR-100 80/20-25}

\paragraph{SNA module.}

\begin{figure}[t]
  \centering
    \includegraphics[width=\linewidth]{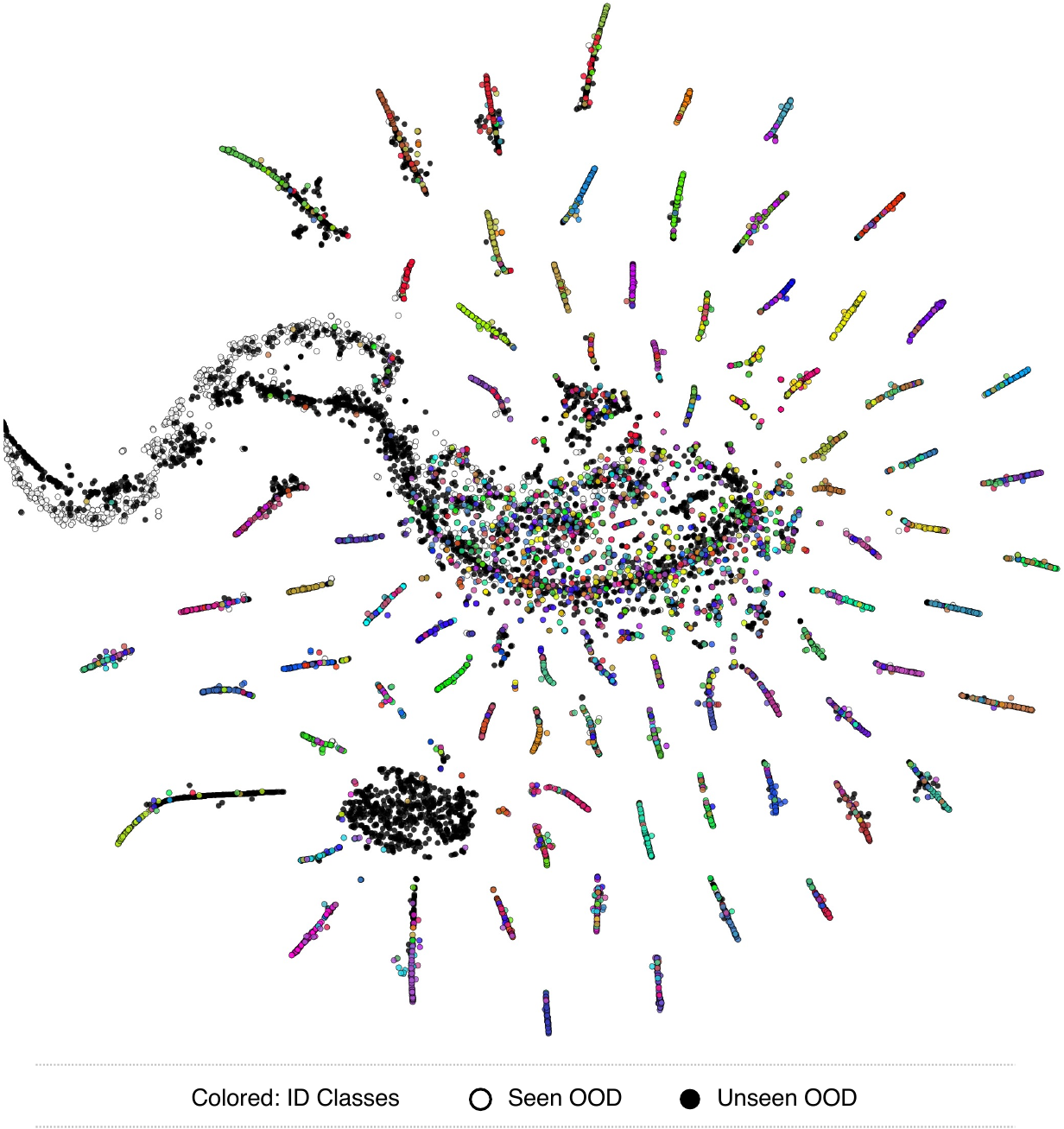}
    \caption{t-SNE visualization of projection embeddings $\mathbf{z}$ of SNA module (CIFAR100 80/20-25).
    }
  \label{fig:sna_cifar100}
\end{figure}

Figure~\ref{fig:sna_cifar100} shows the t-SNE visualization of projection embeddings $\mathbf{z}$ from the SNA module in the CIFAR-100 80/20-25 setting. Despite the increased complexity of the dataset, ID samples form well-separated, elongated clusters extending along distinct directions from the origin. These directional manifolds reflect the angular pull induced by SNA, which consistently aligns confident ID embeddings toward their corresponding class prototypes.

In contrast, OOD samples remain concentrated near the center in a diffuse, unstructured manner. They do not attach to any ID cluster, preserving a clear void between classes—a result of the angular repulsion from all prototypes. Some OOD embeddings stretch along partial directions but never form compact groups.
Seen OOD samples, used as negative labels during training, tend to remain more clearly separated from ID regions. Unseen OODs, lacking such supervision, occasionally intrude into the periphery of ID areas. This geometric structure highlights how SNA achieves strong intra-class alignment, preserves inter-class separation, and maintains a low-density region for OOD samples.

\paragraph{OOD detector.}

\begin{figure}[t]
  \centering
    \includegraphics[width=\linewidth]{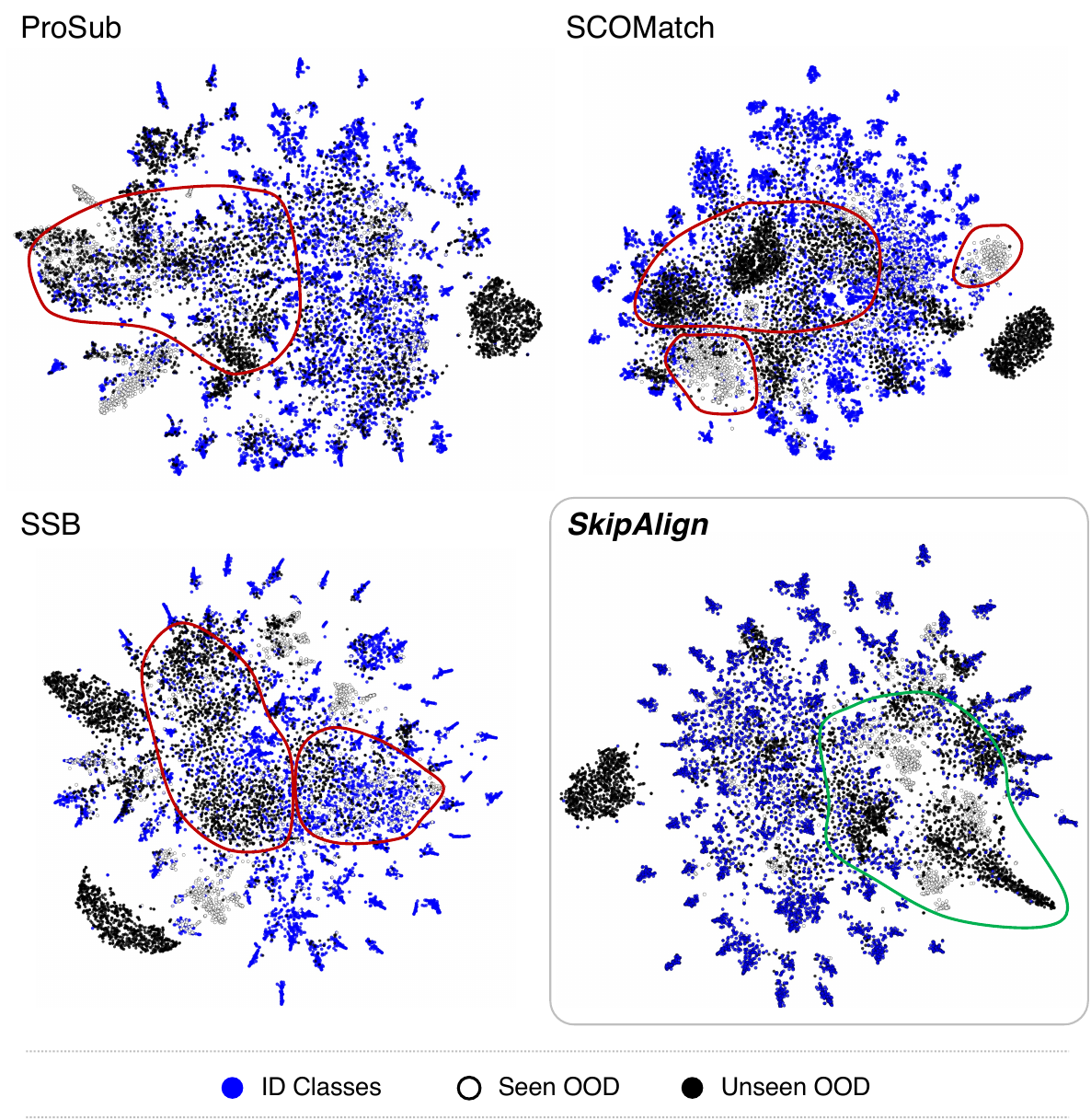}
    \caption{t-SNE visualization of feature embeddings from the OOD detector (CIFAR100 80/20-25).
    }
  \label{fig:ood_cifar100}
\end{figure}

Figure~\ref{fig:ood_cifar100} shows the t-SNE visualization of feature embeddings from the OOD detector on the CIFAR-100 80/20-25 setting. ProSub exhibits ID clusters that are loosely scattered, with substantial overlap from OOD samples, particularly seen OODs, within the ID region. This intrusion likely stems from its broad subspace alignment strategy, which tends to pull ambiguous OOD samples into the ID subspace. SCOMatch fails to form a coherent cluster for seen OODs, especially in the presence of many diverse classes. As a result, its synthetic OOD class does not generalize well, and unseen OODs often go undetected. SSB appears to produce relatively clean ID clusters, but these clusters are small and under-represented due to its conservative use of unlabeled data. Consequently, many ID samples remain mixed with seen and unseen OODs near the center, resulting in fuzzy decision boundaries.
In contrast, \proposal confidently aligns a larger portion of unlabeled ID samples, resulting in larger and more cohesive ID clusters. OOD samples are consistently pushed toward peripheral, low-density regions due to the effect of selective non-alignment, which applies angular repulsion and avoids prototype attraction for uncertain samples. As a result, they remain in the low-density void between ID manifolds and rarely overlap with them. While the central region may appear visually mixed, the ID samples form well-defined clusters, and the sparse distribution of some unseen OODs within this area is clearly distinguishable—unlike SSB, where both ID and OOD are diffusely scattered without clear separation of ID clusters.


\section{D. Ablation Study}

\subsection{D.1 Projection Nonlinearity in SNA}
\begin{table}[t]
\centering
\begingroup
\setlength{\tabcolsep}{2.8mm} 
\small
\renewcommand{\arraystretch}{1.1}
\begin{tabular}{c c ccc}
\toprule
\multirow{2}{*}{\textbf{MLP}} & 
\multirow{2}{*}{\textbf{Acc.}} & 
\multicolumn{3}{c}{\textbf{AUC}} \\
\cmidrule(lr){3-5}
 & & Seen & Unseen & Overall \\
\midrule
           & 71.2 & 89.5 & 81.2 & 82.4 \\
\checkmark & 71.8 & 90.8 & 83.0 & 84.1 \\
\bottomrule
\end{tabular}
\caption{Effect of adding nonlinearity (ReLU) to the projection head of the SNA module (CIFAR-100 55/45-25).}
\label{tab:supple_nonlinear}
\endgroup
\end{table}

To assess the impact of nonlinearity in the projection head of the SNA module, we compare a simple linear projection against a 1-layer MLP with ReLU activation. As shown in Table~\ref{tab:supple_nonlinear}, adding nonlinearity improves both classification accuracy and OOD detection performance across seen and unseen settings.
We attribute this gain to the MLP’s ability to reparameterize the projected embedding space, allowing the angular repulsion signal from SNA to propagate more effectively to the shared backbone. Specifically, nonlinearity introduces a data-dependent Jacobian $\frac{\partial z}{\partial f}$, which enables the angular gradient $\nabla_z \mathcal{L}_{\text{USNA}}$ to be mapped to a wider range of tangential directions in the backbone feature space. This results in a more effective suppression of radial norm growth induced by noisy pseudo-labels. Moreover, the nonlinearity helps to decouple the direction of projected embeddings $z$ from the backbone feature $f$, thereby amplifying the orthogonal component of the SNA gradient. Together, these effects lead to sharper ID clusters and a more diffuse OOD region, enhancing both intra-ID compactness and ID/OOD separability.

\subsection{D.2 Effect of SNA Loss Weight}
\begin{table}[t]
\centering
\begingroup
\setlength{\tabcolsep}{2.8mm} 
\small
\renewcommand{\arraystretch}{1.1}
\begin{tabular}{c c ccc}
\toprule
\multirow{2}{*}{$\lambda_{\text{SNA}}$} & 
\multirow{2}{*}{\textbf{Acc.}} & 
\multicolumn{3}{c}{\textbf{AUC}} \\
\cmidrule(lr){3-5}
 & & Seen & Unseen & Overall \\
\midrule
0.012 & 93.1 & 99.5 & 95.3 & 95.9 \\
0.010 & 93.6 & 99.8 & 95.3 & 95.9 \\
0.008 & 93.6 & 99.5 & 96.2 & 96.7 \\
0.006 & 93.7 & 99.5 & 95.8 & 96.3 \\
0.004 & 93.7 & 99.3 & 95.7 & 96.2 \\
\bottomrule
\end{tabular}
\caption{Ablation study on the loss weight $\lambda_{\text{SNA}}$ of the SNA loss in \proposal (CIFAR10 6/4-50).}
\label{tab:supple_sna_weight}
\endgroup
\end{table}

To assess the sensitivity of our method to the SNA loss weight, we conduct an ablation study by varying $\lambda_{\text{SNA}}$ around the default value of 0.01, which was chosen to match the scale of other loss components, including the closed-set classifier and the OOD detector.
As shown in Table~\ref{tab:supple_sna_weight}, the performance remains consistently high across a wide range of $\lambda_{\text{SNA}}$, demonstrating the robustness of our method to this hyperparameter.
The initial choice of 0.01 already yields strong results, while the best overall performance is observed at $\lambda_{\text{SNA}} = 0.008$, suggesting that this setting provides the best trade-off between classification and OOD detection on the CIFAR-10 6/4-50 setting.

\section{E. Algorithm}

\subsection{E.1 Pseudo-Code}
We present the pseudo-code of \proposal.
The losses related to the Closed-set Classifier follow FixMatch and the losses related to the OOD detector follow OpenMatch and SSB.

\subsection{E.2 Training Details}

For CIFAR-100, we used a 1-layer MLP with ReLU for the closed-set classifier $\mathrm{CC}(\cdot)$ and a 2-layer MLP with ReLU for the outlier detector $\mathrm{OD}(\cdot)$, both with a hidden size of 1024. The projection layer in the SNA module consists of a 1-layer MLP with ReLU, with a hidden size of 256 and a projection size of 128.
For CIFAR-10, $\mathrm{CC}(\cdot)$ uses a smaller hidden size of 128, and the projection layer's hidden size is also reduced to 128.
For ImageNet30 and TinyImageNet, $\mathrm{CC}(\cdot)$ uses a hidden size of 256, and the projection layer uses a hidden size of 256, while the projection size remains 128.
For CIFAR-100, the model was trained with a batch size of 256, $\gamma=2$, and 1024 iterations per epoch for 256 epochs using 4 GPUs (RTX 3090). For CIFAR-10, training was conducted on a single GPU (RTX 3090) with a batch size of 64, $\gamma=4$, over 256 epochs. For ImageNet30, we used 4 GPUs (RTX 3090) with a batch size of 64, $\gamma=2$, for 512 epochs. For TinyImageNet, the model was trained on a single GPU (H100) with a batch size of 32 and $\gamma=6$ for 512 epochs. All experiments used the SGD optimizer with momentum 0.9, and a cosine annealing learning rate schedule starting from 0.03. Weight decay was set to 0.001 for CIFAR-100 and 0.0005 for all other datasets. All experiments were run with three random seeds: 0, 1, and 2. Additional hyperparameter details are available in the experimental configuration files provided in the code repository.
Code is available at \url{https://github.com/yourimchoi/skipalign}

\begin{algorithm*}
\caption{\proposal Pseudo Code}
\label{alg:pseudo_code}
\begin{algorithmic}[1]

\STATE \textbf{Input:} Labeled dataset $\mathcal{D}_l$, labels $y \in \{1,\dots,K\}$, unlabeled dataset $\mathcal{D}_u$,
\STATE \hspace{1em} data augmentations $\mathcal{T}_t$ for $t \in \{w,w',s\}$,
\STATE \hspace{1em} feature extractor(Shared backbone) $\mathcal{F}(\cdot)$, classifiers $g_{CC}$ and OOD detector $g_{OD}$,
\STATE \hspace{1em} output heads $CC(\cdot)$, $OD(\cdot)$, projection head $proj(\cdot)$,
\STATE \hspace{1em} class prototypes $\mu$, iteration count $It$

\FOR{$i = 1$ to $It$}
    \STATE \textit{/* Training Data Preparation */}
    \STATE Sample labeled batch $X = \{(x_i, y_i)\}_{i=1}^{B}$
    \STATE Sample unlabeled batch $\mathcal{U} = \{u_i\}_{i=1}^{\gamma B}$
    \STATE Apply augmentations:
    \STATE \hspace{1em} $x_i^t = \{\mathcal{T}_w(x_i), \mathcal{T}_{w'}(x_i), \mathcal{T}_s(x_i)\}$
    \STATE \hspace{1em} $u_i^t = \{\mathcal{T}_w(u_i), \mathcal{T}_{w'}(u_i), \mathcal{T}_s(u_i)\}$
    \STATE
    \STATE \textit{/* Forward Pass */}
    \STATE $h_i^t = \mathcal{F}(x_i^t)$, \quad $h_{u,i}^t = \mathcal{F}(u_i^t)$
    \STATE $z_i^t = proj(h_i^t)$, \quad $z_{u,i}^t = proj(h_{u,i}^t)$
    \STATE $p_i^t = CC(h_i^t) \in \mathbb{R}^K$, \quad $p_{u,i}^t = CC(h_{u,i}^t)$
    \STATE $\varphi_i^t = OD(h_i^t) \in \mathbb{R}^{2K} \quad \triangleright\ \varphi_{i,k}^t = \{\varphi_{i\text{-}ID,k}^t,\ \varphi_{i\text{-}OOD,k}^t\}$
    \STATE $\varphi_{u,i}^t = OD(h_{u,i}^t) \in \mathbb{R}^{2K}$
    
    \STATE
    \STATE \textit{/* Compute Losses */}
    \STATE $\mathcal{L}_{USNA} = \text{Unlabeled\_SNA\_Loss}(z_{u,i}^w) \quad \triangleright \text{ Eq.}~\ref{eqn:eq2}$
    \STATE $\mathcal{L}_{IA} = \text{Instance-wise\_Alignment\_Loss}(z_i^w) \quad \triangleright \text{ Eq.}~\ref{eqn:eq3}$
    \STATE $\mathcal{L}_{PA} = \text{Prototype-based\_Alignment\_Loss}(z_i^w) \quad \triangleright \text{ Eq.}~\ref{eqn:eq4}$
    \STATE $\mathcal{L}_{SNA}=\lambda_{USNA}\mathcal{L}_{USNA}(\mathcal{U})+\lambda_{IA}\mathcal{L}_{IA}(X)+\lambda_{PA}\mathcal{L}_{PA}(X)$
    
    \STATE $\mathcal{L}_{CC} = \mathcal{L}_x(X) +\lambda_u \mathcal{L}_u(\mathcal{U})$
    \STATE $\mathcal{L}_{OD} = \mathcal{L}_{od}^{OVA}(X) + \lambda_{em} \mathcal{L}_{od}^{em}(\mathcal{U}) + \lambda_{SOCR} \mathcal{L}_{od}^{SOCR}(\mathcal{U})$ + $\lambda_{neg} \mathcal{L}_{od}^{neg}(\mathcal{U})$
    \STATE \quad $\mathcal{L}_{od}^{OVA}(\mathcal{X})=\sum_{i=1}^B\sum_{k=1}^K- \mathds{1}(y_i = k) \log \sigma(s_k^{\text{ID}})- \mathds{1}(y_i \neq k) \log \sigma(s_k^{\text{OOD}})$
    \STATE \quad $\mathcal{L}_{od}^{em}(\mathcal{U})=-\frac{1}{\gamma B}\sum_{i=1}^{\gamma B}\sum_{k=1}^{K}\{\varphi_{i,k}^{ID}\log(\varphi_{i,k}^{ID})+\varphi_{i,k}^{OOD}\log(\varphi_{i,k}^{OOD})\}$
    \STATE \quad $\mathcal{L}_{od}^{SOCR}(\mathcal{U})= \frac{1}{\gamma B}\sum_{i=1}^{\gamma B}\sum_{k=1}^{K}|s_{i,k,\mathcal{T}_w}^{ID}-s_{i,k,\mathcal{T}_{w'}}^{ID}|^2 $
    \STATE \quad $\mathcal{L}_{od}^{neg}(\mathcal{U})=-\frac{1}{\gamma B}\sum_{i=1}^{\gamma B}\frac{1}{\sum_{k=1}^K\mathds{1}(\varphi_{i,k}^{ID}<\eta_{neg})}\sum_{k=1}^K\mathds{1}(\varphi_{i,k}^{ID}<\eta_{neg})\log(1-\varphi_{i,k}^{ID})$

    \STATE $\mathcal{L}_{Total}=\lambda_{CC}\mathcal{L}_{CC}+\lambda_{OD}\mathcal{L}_{OD}+\lambda_{SNA}\mathcal{L}_{SNA}$

    \STATE Update model with loss $\mathcal{L}$

    \IF{current iteration is a prototype update step}
        \STATE Call \textbf{Prototype\_generation}($z_i^w, z_{u,i}^w$)
    \ENDIF
\ENDFOR

\STATE
\STATE \textbf{Prototype\_generation($z_i^w$, $z_{u,i}^w$):}
\FOR{$k = 1$ to $K$}
    \STATE $\mu_{lk} = \text{mean}(z_i^w \mid y_i = k)$
    \STATE $\mu_{uk} = \text{mean}(z_{u,i}^w \mid \text{reliable ID for class } k)$
    \STATE $\mu_k = w_{lk}^{norm} \cdot \mu_{lk} + w_{uk}^{norm} \cdot \mu_{uk} \quad \triangleright \text{ Eq.}~\ref{eqn:eq19}$
\ENDFOR

\end{algorithmic}
\end{algorithm*}

\clearpage

\end{document}